%% file: main.tex
\documentclass{article}

\usepackage[final, nonatbib]{neurips_2021}

\usepackage{lipsum}

\usepackage[utf8]{inputenc} 
\usepackage[T1]{fontenc}    
\usepackage{tgtermes}

\usepackage{url}            
\usepackage{booktabs}       
\usepackage{amsfonts}       
\usepackage{nicefrac}       
\usepackage{microtype}      

\usepackage[english]{babel}

\usepackage{amsmath}
\usepackage{enumitem}
\usepackage{amssymb}
\usepackage{graphicx}
\usepackage{bm}
\usepackage{theorem}
\usepackage[colorlinks=true, allcolors=blue]{hyperref}

\usepackage{mathtools}

\usepackage[title]{appendix}
\usepackage{comment}
\usepackage{amsfonts}
\usepackage{dsfont}
\usepackage{hyperref}
\usepackage{lipsum}
\usepackage{dsfont,subcaption,float}
\usepackage{algorithm}
\usepackage{algorithmic}

\usepackage{thm-restate}
\usepackage[size=small,labelfont=bf]{caption}

\usepackage{xcolor}

\input{macros}

\newcommand{\Epsq}{\Eps_{l,4}}
\newcommand{\Epst}{\Eps_{l,3}}
\newcommand{\Epsd}{\Eps_{l,2}}
\newcommand{\mXt}{\mX_{l,3}}

\title{Reinforcement Learning in Reward-Mixing MDPs}

\author{%
  Jeongyeol Kwon \\
  The University of Texas at Austin \\
  \texttt{kwonchungli@utexas.edu} \\
  \And
  Yonathan Efroni \\
  Microsoft Research, NYC \\
  \texttt{jonathan.efroni@gmail.com} \\
  \AND
  Constantine Caramanis \\
  The University of Texas at Austin  \\
  \texttt{constantine@utexas.edu} \\
  \And
  Shie Mannor \\
  Technion, NVIDIA \\
  \texttt{shie@ee.technion.ac.il, } \\
  \texttt{smannor@nvidia.com}
}

\begin{document}
\maketitle


\begin{abstract}
    Learning a near optimal policy in a partially observable system remains an elusive challenge in contemporary reinforcement learning. In this work, we consider episodic reinforcement learning in a reward-mixing Markov decision process (MDP). There, a reward function is drawn from one of multiple possible reward models at the beginning of every episode, but the identity of the chosen reward model is not revealed to the agent. Hence, the latent state space, for which the dynamics are Markovian, is not given to the agent. We study the problem of learning a near optimal policy for two reward-mixing MDPs. Unlike existing approaches that rely on strong assumptions on the dynamics, we make no assumptions and study the problem in full generality. Indeed, with no further assumptions, even for two switching reward-models, the problem requires several new ideas beyond existing algorithmic and analysis techniques for efficient exploration. We provide the first polynomial-time algorithm that finds an $\epsilon$-optimal policy after exploring $\tilde{O}(poly(H,\epsilon^{-1}) \cdot S^2 A^2)$ episodes, where $H$ is time-horizon and $S, A$ are the number of states and actions respectively. This is the first efficient algorithm that does not require any assumptions in partially observed environments where the observation space is smaller than the latent state space. 
\end{abstract}

\section{Introduction}
\input{Introduction}

\section{Problem Setup}
\input{Preliminaries}

\section{Algorithms}
\label{section:main_algorithm}
\input{Algorithm}

\section{Analysis Overview}
\label{section:main_analysis}
\input{Analysis}


\section{Discussions and Future Work}
\label{subsection:main_discussion}
We developed the first efficient algorithm for RL in two reward-mixing MDPs without any assumptions on system dynamics. We conclude this work with discussions on our assumptions and several future directions. 

\paragraph{Non-uniform/Unknown Mixing Weights.} When the prior over contexts is not uniform, {\it i.e.,} $w \neq 1/2$, parameter recovery procedure (Algorithm \ref{algo:LMDP_recovery}) may be more complicated due to the sign-ambiguity issue in $p_-(x)$. When $w = 1/2$, the estimated model is identical whether we use $\hat{p}_-(x) \approx p_-(x)$ or $\hat{p}_-(x) \approx -p_-(x)$. However, the model is no longer identical when we have correct signs of $p_-(x)$ or not if $w \neq 1/2$. In Appendix \ref{appendix:sign_ambiguity}, we discuss the extension to non-uniform mixing weights in a great detail. Once we can solve with any known $w \in [0,1]$, then one possible solution to handle unknown mixing weight is to sweep all possible mixing weight for $w$ from 0 to 1 with discretization level $O(\epsilon / H)$, recover an empirical model obtained by solving an LP assuming known $w$ for all discretized weights and pick the best policy by testing each policy on the environment.

\paragraph{Structured Reward Distributions.} Our approach can be extended to structured reward distributions, {\it e.g.,} distributions over finite supports. For instance, if a reward can take $l$ different values in $\{q_1, q_2, ..., q_l\}$, we redefine $\mu(x_1, x_2) := \frac{1}{2} \sum_{m=1}^2 p_m(x_1) p_m(x_2)^\top$ as a $l \times l$ moment-matrix with setting $p_m(x) = [R_m(r=q_1|x), R_m(r=q_2|x), ..., R_m(r=q_l|x)]^\top$. Then, we can extend an LP in Algorithm \ref{algo:LMDP_recovery} to find $\hat{p}_m(x)$ to match $\mu(x_1, x_2)$ within confidence intervals. This approach would also pay extra $poly(l)$-factors both in sample and computational complexity. Then, once we can verify that the recovered difference in reward models is close, {\it i.e.,} $\|\hat{p}_-(x) - p_-(x)\|_1 \le O(\epsilon_l)$, then the remaining proof for Theorem \ref{theorem:upper_bound} would follow similar to the Bernoulli reward case. We leave a more thorough investigation in this direction as future work.

\paragraph{Extensions to More Than Two Reward Models $M > 2$.}
As mentioned earlier, RM-MDP with $M$ reward models is an MDP analogy to learning a mixture of $M$ product distributions over reward distributions. If {\it every higher-order correlation of interest between multiple state-actions can be estimated}, we may apply existing methods for learning a mixture of product distributions over a binary field \cite{freund1999estimating, feldman2008learning, jain2014learning}. For instance, Jain and Oh \cite{jain2014learning} proposed a spectral method which requires all off-diagonal elements of up to third order moments to learn parameters of $M$-mixtures of $n$-product distributions. To apply their method, we may estimate correlations of all three different state-actions as similarly in Algorithm \ref{algo:pure_explore}, then reward model recovery of RM-MDP can be converted to learning a mixture of $n=SA$-product reward distributions. 
However, as we have already seen in this paper, the real challenge is in recovering the model from correlations of {\it unevenly} reachable sets of state-actions, and then analyzing errors with different levels of uncertainty in correlations. We believe this is a very interesting direction for future research.

In addition to the above mentioned issues, we believe there are a number of possible future directions. While we specifically focused on RM-MDP problems, the more general problem of Latent MDP \cite{kwon2021rl} has not been resolved even for $M = 2$ without any assumptions. We believe that it would be an interesting future direction to develop efficient algorithms for Latent MDPs with small $M$. We can also consider more technical questions in RM-MDPs, {\it e.g.,} what would be the optimal sample-complexity dependency on $H, \epsilon$, and whether we can develop a fully online algorithm to minimize regret. We leave them as future work.

\subsection*{Acknowledgement}
The research was funded by NSF grant 2019844, the ISF under contract 2199/20, and by the Army Research Office and was accomplished under Cooperative Agreement Number W911NF-19-2-0333. The views and conclusions contained in this document are those of the authors and should not be interpreted as representing the official policies, either expressed or implied, of the Army Research Office or the U.S. Government. The U.S. Government is authorized to reproduce and distribute reprints for Government purposes notwithstanding any copyright notation herein.

We also would like to thank Kurtland Chua for pointing out errors in the Appendix.

\bibliographystyle{abbrv}
\bibliography{main}

\newpage 

\appendix

\begin{appendices}
\input{Appendix}

\end{appendices}

\end{document}

%% file: macros.tex
\newtheorem{theorem}{Theorem}[section]

\newtheorem{corollary}{Corollary}
\newtheorem{lemma}[theorem]{Lemma}

\newtheorem{definition}{Definition}

\newenvironment{proof}{\paragraph{\it Proof.}}{\hfill$\square$}

\newcommand{\aR}{R_+}	

\newcommand{\wQ}{\widetilde{Q}}	
\newcommand{\wpi}{\widetilde{\pi}}	

\newcommand{\mS}{\mathcal{S}}	
\newcommand{\mA}{\mathcal{A}}	
\newcommand{\mT}{\mathcal{T}}	
\newcommand{\mM}{\mathcal{M}}	

\newcommand{\tmM}{\widetilde{\mathcal{M}}}	
\newcommand{\tmS}{\widetilde{\mathcal{S}}}
\newcommand{\tmA}{\widetilde{\mathcal{A}}}

\newcommand{\mX}{\mathcal{X}}	
\newcommand{\indic}[1]{\mathds{1}\left\{ #1 \right\}} 
\newcommand{\one}{\mathbf{1}}

\newcommand{\PP}{\mathds{P}}    

\newcommand{\Eps}{\mathcal{E}}

\newcommand{\Exs}{\mathbb{E}}

\allowdisplaybreaks

\newif\ifdraft
\drafttrue  

%% file: Introduction.tex
In reinforcement learning (RL), an agent solves a sequential decision-making problem in an unknown dynamic environment to maximize the long-term reward \cite{sutton2018reinforcement}. The agent interacts with the environment by receiving feedback on its actions in the form of a state-dependent reward and observation. 

One of the key challenges in RL is exploration: in the absence of auto-exploratory properties such as ergodicity of the system, the agent has to devise a clever scheme to collect data from under-explored parts of the system. In a long line of work, efficient exploration in RL has been extensively studied for fully observable environments, {\it i.e.,} under the framework of Markov decision process (MDP) with a number of polynomial-time algorithms proposed \cite{kearns2002near, jaksch2010near, russo2013eluder, osband2014model, efroni2019tight}. For instance, in tabular MDPs, {\it i.e.,} environments with a finite number of states and actions, we can achieve sample-optimality or minimax regret without any assumptions on system dynamics \cite{azar2017minimax, zanette2019tighter}. 

In contrast to fully observable environments, very little is known about the exploration in partially observable MDPs (POMDPs). In general, RL in POMDPs may require an exponential number of samples (without simplifying structural assumptions) \cite{krishnamurthy2016pac, jin2020reward}. Therefore, it is important to consider natural sub-classes of POMDPs which admit tractable solutions. Previous studies focus on a special class of POMDPs where observation spaces are large enough such that a single observation provides sufficient information on the underlying state \cite{guo2016pac, krishnamurthy2016pac, azizzadenesheli2016reinforcement, dann2018oracle, du2019provably, jin2020sample}. However, there are many applications in robotics, topic modeling and beyond, where the observations space is smaller than the latent state space. This ``small observation space'' setting is critical, as most prior work does not apply. Our work seeks to provide one of the first results to tackle a class of problems in this space.


In particular, we tackle a sub-class of POMDPs inspired by latent variable or mixture problems. Specifically, we consider reward-mixing MDPs (RM-MDPs). In RM-MDPs, the transition kernel and initial state distribution are defined as in standard MDPs, while the reward function is randomly chosen from one of $M$-reward models at the beginning of every episode (see the formal definition \ref{definition:rm_lmdp}). RM-MDPs are important in their own right. Consider, for instance, a user-interaction model in a dynamical web system, where reward models vary across users with different characteristics \cite{maillard2014latent, hallak2015contextual}. A similar setting has been considered in a number of recent papers \cite{chades2012momdps,brunskill2013sample,hallak2015contextual,steimle2018multi,buchholz2019computation}, and most recently, and most directly related to our setting, \cite{kwon2021rl}. 

Even for $M=2$, the RM-MDP setting shares some of the key challenges of general POMDPs. This is because the best policy for the RM-MDP problem should account for not just the current state, but also an entire sequence of previous observations since some previous events might be correlated with a reward of the current action. And in the small observation space setting, prior work as in \cite{guo2016pac, krishnamurthy2016pac, azizzadenesheli2016reinforcement, dann2018oracle, du2019provably, jin2020sample} cannot be applied. The work in \cite{kwon2021rl} develops an algorithmic framework for solving the RM-MDP problem in this small observation setting, but requires strong assumptions without which the algorithm falls apart. We discuss this in more detail below. Indeed, no work to date, has been able to address the RM-MDP problem, even for $M=2$, without significant further assumptions. This is precisely the problem we tackle in this paper.

\paragraph{Main Results and Contributions} 
We focus on the problem of learning near-optimal policies in two reward-mixing MDPs, {\it i.e.,} RM-MDPs with two reward models $M = 2$. To the best of our knowledge, no prior work has studied sample complexity of RL in RM-MDPs even for $M = 2$ without any assumptions on system dynamics. Specifically, we provide the first polynomial-time algorithm which learns an $\epsilon$-optimal policy after exploring $\tilde{O} (poly(H, \epsilon^{-1}) \cdot S^2 A^2)$ episodes. We also show that $\Omega(S^2 A^2 / \epsilon^2)$ number of episodes is necessary, and thus our result is tight in $S, A$. This is the first efficient algorithm in a sub-class of partially observable domains with small observation spaces and without any assumptions on system dynamics.

On the technical side, we must overcome several new challenges that are not present in fully observable settings. One key hurdle relates to identifiability. In standard MDPs, an average of single observations from a single state-action pair is sufficient to get unbiased estimators of transition and reward models for the state-action. However, in RM-MDPs, the same quantity would only give an \emph{averaged reward} model, where the average is taken over the reward model distribution. However, an average reward model alone is not sufficient to obtain the optimal policy which depends on a sequence of previous rewards. 

Our technique appeals to the idea of {\it uncertainty in higher-order moments}, bringing inspiration from algorithms for learning a mixture of structured distributions \cite{feldman2008learning, chan2013learning, jain2014learning, diakonikolas2019robust}. In such problems, the key for efficient learning is to leverage information from higher-order moments. Following this spirit, we consider estimating correlations of rewards at multiple different state-actions. A central challenge we must overcome is that in finite-horizon MDPs, it may be not possible to estimate all higher-order correlations with uniformly good accuracy. In fact, it may not be possible to estimate some correlations at all when some pairs of state-actions are not reachable in the same episode. Therefore, we cannot simply rely on existing techniques that require good estimations of {\it all} elements in higher-order moment matrices. The main technical challenge is, therefore, to show that a near-optimal policy can still be obtained from {\it uncertain and partial correlations} of state-actions. Our technical contributions are thus two-fold: (1) design of an efficient exploration algorithm to estimate each correlation up to some required accuracy, and (2) new technical tools to analyze errors from different levels of uncertainties in estimated higher-order correlations.

\paragraph{Related Work} Efficient exploration in fully observed environments has been extensively studied in the framework of MDPs. In tabular MDPs, a long line of work has studied efficient exploration algorithms and their sample complexity based on the principle of optimism \cite{jaksch2010near, russo2013eluder, azar2017minimax, efroni2019tight, zanette2019tighter, tossou2019near, simchowitz2019non} or posterior-sampling \cite{asmuth2009bayesian, russo2013eluder, osband2014model, osband2017posterior}. Beyond tabular settings, recent work seeks for efficient exploration in more challenging settings such as with large or continuous state spaces \cite{jiang2017contextual, jin2020provably, du2019good}. All aforementioned work considers only fully observable environments where the complexity of exploration is significantly lower as there is no need to consider observation histories.

Most previous study on exploration in POMDPs focuses on large observation spaces such that a set of single observations is statistically sufficient to infer internal states, {\it i.e.,} the observation matrix has non-zero minimum singular value. Under such environments, previous work either considers a strong assumption on the uniform ergodicity of the latent state transitions \cite{hsu2012spectral, boots2011closing, guo2016pac, azizzadenesheli2016reinforcement} or the uniqueness of underlying states for every observation \cite{krishnamurthy2016pac, dann2018oracle, du2019provably}. The exception is a recent work by Jin et al. \cite{jin2020sample} which studied efficient exploration without any assumptions on latent system dynamics. 

One of the most closely related to problem to our work is a reinforcement learning problem with latent contexts (also referred as multitask RL), which is considered in \cite{brunskill2013sample, hallak2015contextual, kwon2021rl}. Previous approaches relied on several strong assumptions such as very long time horizon \cite{brunskill2013sample, hallak2015contextual}, revealed contexts in hindsight or ergodic properties of system dynamics \cite{kwon2021rl}. Our work takes a first step without such assumptions by focusing on environments where only the reward model differs across different tasks. 

RM-MDP might be also viewed as a special case of adversarial MDPs ({\it e.g.,} \cite{yu2009markov, neu2012adversarial, jin2019learning}) with an oblivious adversary who can only play within a finite set of reward models. However, our goal is to find an optimal policy in a broader class of {\it history-dependent} policies, whereas in adversarial MDP literature they compare only to the best Markovian policy in hindsight. RM-MDP can be also considered as an instance of latent variable models ({\it e.g.,} \cite{de1989mixtures, freund1999estimating, dasgupta1999learning, hsu2012spectral, hoffmann2020learning}), where sample reward sequences are influenced by unobserved confounders. RM-MDP poses several new challenges as both challenges from reinforcement learning and statistical inference are interlaced by latent contexts.


%% file: Preliminaries.tex
We define the problem of episodic reinforcement learning problem with time-horizon $H$ in reward-mixing Markov decision processes (RM-MDPs) defined as follows: 
\begin{definition}[Reward-Mixing Markov Decision Process (RM-MDP)]
    \label{definition:rm_lmdp}
    Let $\mathcal{M}$ be a tuple $(\mS, \mA, T, \nu, \{w_m\}_{m=1}^M, \{R_m\}_{m=1}^M)$ be a RM-MDP on a state space $\mathcal{S}$ and action space $\mathcal{A}$ where $T: \mathcal{S}\times\mathcal{A}\times\mathcal{S} \rightarrow [0,1]$ is a common transition probability measures that maps a state-action pair and a next state to a probability, and $\nu$ is a common initial state distribution. Let $S = |\mS|$ and $A = |\mA|$. $w_1, ..., w_M$ are the mixing weights such that at the beginning of every episode, one reward model $R_m$ is randomly chosen with probability $w_m$. A reward model $R_m: \mathcal{S}\times\mathcal{A} \times \{0,1\} \rightarrow [0, 1]$ is a probability measure for rewards that maps a state-action pair and a binary reward to a probability for $m \in [M]$. 
\end{definition}
In this work, we focus on a special case of RM-MDPs when $M = 2$, {\it i.e.,} 2RM-MDPs. For the ease of presentation, we assume all random rewards take values from $\{0,1\}$, {\it i.e.,} rewards are binary random variables. In this work, we assume a uniform-prior over each latent context such that $w_1 = w$ and $w_2 = 1 - w$ with $w = 1/2$. Detailed discussions on these simplifying assumptions can be found in Section \ref{subsection:main_discussion}. We consider a policy class $\Pi$ which contains all history-dependent policies $\pi: (\mS, \mA, \{0,1\})^* \times \mS \rightarrow \mA$. The goal of the problem is to find a near optimal policy $\pi \in \Pi$ that has near optimal value w.r.t. the optimal one, $V_{\mM}^* := \max_{\pi \in \Pi} \sum_{m=1}^M w_m \Exs_{m}^\pi \left[ \sum_{t=1}^H r_t \right],$ where $\Exs_{m}^\pi [\cdot]$ is expectation taken over the $m^{th}$ MDP with a policy $\pi$. More formally, we wish our algorithm to return an $(\epsilon,\eta)$ provably-approximately-correct (PAC) optimal policy, which we also refer as a near optimal policy, defined as follows:
\begin{definition}[$(\epsilon,\eta)$-PAC optimal policy]\label{definition: pac optimal policy}
    An algorithm is $(\epsilon,\eta)$-PAC if it returns a policy $\hat{\pi}$ s.t.
    \begin{align*}
        \PP( V_{\mM}^* - V_{\mM}^{\hat{\pi}} \le \epsilon ) \ge 1 - \eta.
    \end{align*}
\end{definition}

\subsection{Notation}
We often denote a state-action pair $(s,a)$ as one symbol $x = (s,a) \in \mS \times \mA$. For any sequence $(y_1, y_2, ..., y_t)$ of length $t$, we often simplify the notation as $(y)_{1:t}$ for any symbol $y$. We denote $l_1$ sum of any function $f$ over a random variable $X$ conditioned on an event $E$ as $\| f (X | E) \|_1 = \sum_{X \in \mX} |f (X | E)|$, where $\mX$ is a support of $X$. $\indic{\Eps}$ is an indicator function for any event $\Eps$ and $\PP(\Eps)$ is a probability of any event $\Eps$ measured without contexts. If we use $\PP$ without any subscript, it is a probability of an event measured outside of the context, {\it i.e.}, $\PP(\cdot) = \sum_{m=1}^M w_m \PP_m(\cdot)$. We refer $\PP_m$ the probability of any event measured in the $m^{th}$ context (or in $m^{th}$ MDP). If probability of an event depends on a policy $\pi$, we add superscript $\pi$ to $\PP$. We denote $V_\mM^\pi$ as an expected long-term reward for model $\mM$ with policy $\pi$. We use $\hat{\cdot}$ to denote empirical counterparts. For any set $\mA$ and $d \in \mathbb{N}$, $\mA^{\bigotimes d}$ is  a $d$-ary Cartesian product over $\mA$. We use $a \vee b$ to refer $\max(a,b)$ and $a \wedge b$ to mean $\min(a,b)$ for $a,b \in \mathbb{R}$. 

Specifically for $M=2$: We use shorthand $p_m(x) := R_m(r = 1|x)$ for $m = 1,2$.  Let an expected averaged reward $p_+ (x) := \frac{1}{2} (p_1(x) + p_2(x))$, differences in rewards $p_- (x) := \frac{1}{2} (p_1(x) - p_2(x))$ and $\Delta(x):= |p_-(x)|$. 


%% file: Algorithm.tex
\begin{algorithm}[t]
    \caption{Learning Two Reward-Mixture MDPs}
    \label{algo:two_reward_mixture}
    
    \begin{algorithmic}[1]
        \STATE {Run pure-exploration (Algorithm \ref{algo:pure_explore}) to estimate second-order correlations}
        \STATE {Estimate 2RM-MDP parameters $\hat{\mM}$ from the collected data (Algorithm \ref{algo:LMDP_recovery})}
        \STATE {Return $\hat{\pi}$, the (approximately) optimal policy of $\hat{\mM}$}
    \end{algorithmic}
\end{algorithm}

Before developing a learning algorithm for 2RM-MDP, let us provide intuition behind our algorithm. At a high-level, our approach lies on the following observation: 
\begin{center}
    \emph{The latent reward model of 2RM-MDP can be recovered from reward \emph{correlations} and the \emph{averaged} reward}.
\end{center}
Consider a simpler interaction model, in which we have a perfect and exact access to correlations of the reward function. That is,  suppose we can query for any $(x_i,x_j) \in (\mS \times \mA)^{\bigotimes 2}$ the reward correlation function
\begin{align}
    \mu(x_i, x_j) &:= \Exs \left[r_i \cdot r_j | x_i, x_j \right] = \frac{1}{2} \cdot (p_1 (x_i)p_1(x_j) + p_2 (x_i)p_2(x_j)). \label{eq:def_mu}
\end{align}
Assume we can also query the exact expected average reward $p_+(x)$ for all $x \in \mS \times \mA$. Suppose we can construct a matrix $B \in \mathbb{R}^{SA \times SA}$ indexed by state-actions $x \in \mS \times \mA$, such that at its $(i,j)$ entry is given by:
\begin{align*}
    B_{i,j} = \mu(x_i, x_j) - p_+(x_i) p_+(x_j).
\end{align*}
Simple algebra shows that $B = q q^\top$ where $q \in \mathbb{R}^{SA}$ is a vector indexed by $x \in \mS \times \mA$ such that $q_i = p_-(x_i)$. Hence, we can compute the top principal component of $B$, from which we can recover $p_-(x)$ for all $x\in \mS \times \mA$. Through the access to $p_-(x)$ (ignore sign ambiguity issue for now) and $p_+(x)$ we can recover the probabilities conditioned on the latent contexts via
\begin{align}
    R_1(r=1|x) = p_+(x) + p_-(x) \ \textit{ and } \ R_2(r=1|x) = p_+(x) - p_-(x). \label{eq: recovery of reward}
\end{align}
Once we have the latent reward model we can use any approximate planning algorithm (e.g., point-based value iteration) to find an $\epsilon$-optimal policy.


However, when such exact oracle is not supplied, and the interaction with the 2RM-MDP is based on trajectories and roll-in policies, getting a point-wise good estimate of $B$ is not generally possible. For example, some state-action pairs cannot be visited in the same episode, in which case we cannot get any samples for the correlations of such pairs of state-actions: $\mu(x_i,x_j)$ is completely unknown. In such a case, recovery of the true model parameters is not possible in general even if we estimate all reachable pairs {\it without errors}. Hence the main challenge is as follows: can we estimate a model from a set of empirical second-order correlations such that the estimated model is close to the true model in terms of the optimal policy, even if it is not close in model parameters?

For usual MDPs, states that cannot be reached pose no problems: we do not need to learn transition or reward probabilities of these states, since no optimal policy will use that state. Indeed, with this observation at hand, {\it reward free exploration} techniques~\cite{jin2020reward, kaufmann2021adaptive}  can be utilized to learn a good model w.r.t. all possible reward functions (given a fixed initial distribution). Our main conceptual contribution is to show how these techniques can be leveraged for the RM-MDP problem.

\paragraph{Overview of our approach.}


Our approach develops the following ideas. 
\begin{enumerate}
\item Section \ref{subsec:pure_exploration}:  We define what we call a {\em Second Order MDP}, an augmented MDP whose states are pairs of states of the original MDP. Borrowing technology from single MDPs, we show that reward free exploration on the augmented MDP  can be used to approximate second order moments of the original MDP, along with confidence intervals, obtained by how often a particular pair of states can be reached.
\item Section \ref{subsec:modelrecovery}: We show how to find the parameters of a model whose second order moments are in the confidence set obtained from the augmented MDP. To do this, we show that we can use linear programming to select an element of the uncertainty set that corresponds to valid second order moments for a model. 
\item Section \ref{subsec:mainresults}: The main contribution of the analysis, is then to show that for any model whose true second order statistics lie in this set obtained, an approximate planning oracle is guaranteed to return an $\epsilon$-optimal solution to the true model. Theorem \ref{theorem:upper_bound} shows that sample complexity depends quadratically in $(S\cdot A)$. This is well-expected, given that our algorithm must compute statistics on the augmented graph. We next give a matching lower bound: Theorem \ref{theorem:lower_bound} shows that this quadratic dependence is unavoidable.  
\end{enumerate}

\subsection{Pure Exploration of Second-Order Correlations}
\label{subsec:pure_exploration}
Our first goal is to collect samples for \emph{pairs} of state-actions through exploration. This will ultimately allow us to estimate the correlations of the reward function, as well as the average reward per-state $p_+(x)$. The backbone of pure-exploration is adaptive reward-free exploration schemes studied in standard MDP settings \cite{kaufmann2021adaptive}. We first consider a surrogate model that helps to formulate data collection process for correlations of pairs. Specifically, we first define the augmented MDP:
\begin{definition}[Augmented Second-Order MDPs]
    An augmented second-order MDP $\tmM$ is defined on a state-space $\tmS$ and action-space $\tmA$ where 
    \begin{align*}
        \tmS &= \left\{ (i, v, s)| \ i \in \{1,2,3\} , v: (v^1, v^2) \in ((\mS \times \mA) \cup \{null\})^{\bigotimes 2}, s \in \mS \right\}, \\
        \tmA &= \{ (a, z)| \ a \in \mA, z \in \{0,1\} \}.
    \end{align*}
    Then in $\tmM$, an augmented state $\left( i_t, v_t, s_t \right)$ evolves under an action $(a_t, z_t)$ as follows:
    \begin{align}
        &i_1 = 1, v_1 = (null, null), s_1 \sim \nu(\cdot), \quad s_{t+1} \sim T(\cdot | s_t,a_t), \quad v^j_{t+1} = v^j_t \ \ \forall j \in [2] / \{i_t\}, \nonumber \\
        &i_{t+1} = \begin{cases} i_t & \text{if } z_t = 0 \text{ or } i_t = 3 \\ i_{t} + 1 & \text{else } \end{cases}, \quad v^{i_t}_{t+1} = \begin{cases} v^{i_t}_t & \text{if } z_t = 0 \text{ or } i_t = 3 \\ (s_t, a_t) & \text{else } \end{cases} . \label{eq:augment_rule}
    \end{align}
\end{definition}
The second-order augmented MDP consists of an extended state space where the first two coordinates $i, v$ store previously selected state-actions followed by the coordinate $s$ for the current state. The action space is a product of original actions $a$ and choice actions $z$, where $z = 1$ means that we select the current state-action to be explored as a part of a pair in the episode. Specifically, when $z = 1$, by design we increase the count $i$. If $i$ reaches 3, it means we collected a sample of a pair $(v^1, v^2)$ in the episode. 

\begin{algorithm}[t]
    \caption{Pure Exploration of Second-Order Correlations}
    \label{algo:pure_explore}
    \begin{algorithmic}[1]
        \STATE{Initialize $\widetilde{Q}_{(\cdot)} (\cdot) = \widetilde{V}_0 = 1$, $n(v) = 0$ for all $v \in (\mS \times \mA)^{\bigotimes 2}$.}
        \WHILE{$\widetilde{V}_0 > \epsilon_{pe}$}
            \STATE {Get an initial state $s_1$ for the $k^{th}$ episode. Let $v_1 = (null, null)$, $i = 1$, $r_{c} = 1$}
            \FOR{$t = 1, 2, ..., H$}
                \STATE {Pick $(a_t, z_t) = arg \max_{(a,z) \in \tmA} \widetilde{Q}_t ((i_t, v_t, s_t), (a,z))$.}
                \STATE{Play action $a_t$, observe next state $s_{t+1}$ and reward $r_t$.}
                \IF {$i_t \le 2$ \AND $z = 1$}
                    \STATE{$r_c \leftarrow r_c \cdot r_t$}
                \ENDIF
                \STATE{Update $(i_{t+1}, v_{t+1}, s_{t+1})$ according to the choice of $a_t$ and $z_t$ following the rule in \eqref{eq:augment_rule}}
            \ENDFOR
            \IF{$i_{H+1} = 3$}
                \STATE{$n(v_{H+1}) \leftarrow n(v_{H+1}) + 1$, $\quad \hat{\mu} (v_{H+1}) \leftarrow \left(1 - 1/n(v_{H+1})\right) \hat{\mu}(v_{H+1}) + r_c$}
            \ENDIF
            \STATE{Update $\hat{\nu}, \hat{T}, \hat{p}_+$ from the trajectory $(s,a,r)_{1:H}$, and then update $\widetilde{Q}$ and $\widetilde{V}$ using \eqref{eq:update_q2}, \eqref{eq:optimistic_value}}
        \ENDWHILE
    \end{algorithmic}
\end{algorithm}

Initially the true model parameters are unknown and thus we are not aware of how to collect samples for any pair of state-actions or how much samples we need for each pair. In order to resolve this, we use ideas from reward free exploration for MDPs: we consider the upper confidence error bound function $\widetilde{Q}$ that follows from the Bellman-equation for $\widetilde{\mM}$ with (pure) exploration bonus:
\begin{align}
    \label{eq:update_q2}
    &b_r(i, v, z) = \indic{i = 2 \cap z = 1} \cdot \left(1 \wedge \sqrt{\frac{\iota_2}{n(v')}} \right), \ b_T(s,a) = \left(1 \wedge \sqrt{\frac{\iota_T}{n(s,a)}} \right), \nonumber \\
    &\widetilde{Q}_t ((i,v,s), (a,z)) = 1 \wedge \left(b_r(i,v,z) + \Exs_{s' \sim \hat{T}(\cdot|s,a)} \left[\widetilde{V}_{t+1} (i',v',s') \right] + b_T(s,a) \right),
\end{align}
where $\indic{i=2 \cap z = 1}$ is an indicator of whether to collect samples for correlations between state-actions in $v'$. Here, $i'$ and $v'$ are first and second coordinates of the next state following the transition rule \eqref{eq:augment_rule}, $\iota_2 = O(\log(K/\eta))$, $\iota_T = O(S \log(K/\eta))$ are properly set confidence interval parameters, $K$ is the total number of episodes to be explored and $\wQ_{H+1}(\cdot) = 0$. At every episode, we choose a greedy action with respect to $\widetilde{Q}$ and collect sufficient data for pairs of state-actions by pure exploration on the augmented MDP. We repeat the pure-exploration process until $\widetilde{V}_0$ becomes less than the threshold $\epsilon_{pe}$ where
\begin{align}
    \widetilde{V}_t(i,v,s) = \max_{(a',z') \in \tmA} \widetilde{Q}_t((i, v, s), (a',z')), \quad \widetilde{V}_0 = \sqrt{\iota_\nu / K} + \sum_{s} \hat{\nu} (s) \cdot \widetilde{V}_1(1, v_1, s),  \label{eq:optimistic_value}
\end{align}
and $\iota_\nu = O(S \log(K/\eta))$ is a confidence interval parameter for initial states. The pure-exploration procedure is summarized in Algorithm \ref{algo:pure_explore}. The main purpose of Algorithm \ref{algo:pure_explore} is to balance the amount of samples for correlations.

\subsection{Recovery of Empirical Model with Uncertainty}
\label{subsec:modelrecovery}
Once we gather enough samples to estimate correlations of pairs of state-actions we are assured that $\tilde{V}_0\leq \epsilon_{pe}$. Given this, we describe an efficient algorithm to recover a good estimate of the true 2RM-MDP parameters. We formulate the model recovery problem as a linear program (LP). Recall that for any $x_i, x_j \in (\mS \times \mA)^{\bigotimes 2}$, we can compute the multiplication of differences at two positions $x_i, x_j$: 
\begin{align} 
    \label{eq:def_u}
    u(x_i, x_j) := p_-(x_i) p_-(x_j) &= \mu(x_i, x_j) - p_+(x_i) p_+(x_j).
\end{align}
Ideally with exact oracles for getting values of $u(x_i, x_j)$, we can recover $p_-(x)$ through LP: let $l(x) := \log|p_-(x)|$ for all $x \in \mS \times \mA$. Then we first find a solution for the following LP:
\begin{align*}
    l(x_i) + l(x_j) = \log |\mu(x_i, x_j) - p_+(x_i) p_+(x_j)|, \ \forall x_i, x_j \in \mS \times \mA.
\end{align*}
After solving for $l(x)$, we can find consistent assignments of signs $sign(x)$ such that ${p_-(x) = sign(x) \cdot \exp(l(x))}$, which can be formulated as 2-Satisfiability problem \cite{aspvall1979linear}.

However,, after the pure-exploration phase, we only have estimates of correlations $\hat{\mu}(x_i, x_j)$ and confidence intervals $b(x_i, x_j) := \sqrt{\iota_2/n(x_i, x_j)} + \epsilon_0^2$ where $\epsilon_0 = \epsilon / H^2$ is a small tolerance parameter (as well as estimates of the expected average reward $\hat{p}_+ (x)$ and its confidence interval $b(x) := \sqrt{\iota_2/n(x)}$). Note that $n(x_i, x_j)$ is the number of counts that a pair $(x_i, x_j)$ is counted during pure-exploration phase. Let $\hat{u}(x_i, x_j) := \hat{p}_-(x_i) \hat{p}_-(x_j)$ be an empirical estimate of $u(x_i,x_j)$. We want to recover $\hat{p}_-(x) = sign(x) \cdot \exp(\hat{l}(x))$ such that the following is satisfied:
\begin{align}
    \label{eq:lp_variables}
    |(\hat{\mu}(x_i, x_j) + \hat{p}_+ (x_i) \hat{p}_+(x_j)) - \hat{u} (x_i, x_j)| \le b(x_i, x_j), \ \forall x_i, x_j \in \mS \times \mA.
\end{align}
We present in Appendix \ref{appendix:formulation_lp} the LP formulation to recover $\hat{l}(x)$ and $sign(x)$. The outline of the procedure is stated in Algorithm \ref{algo:LMDP_recovery} and more details are presented in Algorithm \ref{algo:LMDP_recovery_detail}. In order for Algorithm \ref{algo:LMDP_recovery} to succeed, we need the following lemma on the existence of feasible solutions in the LP with high probability:
\begin{lemma}
    \label{lemma:feasible_optimism}
    Suppose that we set the confidence parameters as $\iota_1 = \iota_2 = O(\log(SA/\eta))$ and $\iota_\nu = \iota_T = O(S \log(SA/\eta))$. Then with probability at least $1 - \eta$, Algorithm \ref{algo:LMDP_recovery} returns a model $\hat{\mM}$ that satisfies constraints \eqref{eq:lp_variables} for all $x_i, x_j \in \mS \times \mA$. 
\end{lemma}
Given Lemma \ref{lemma:feasible_optimism}, Algorithm \ref{algo:LMDP_recovery} is able to find at least one feasible solution, i.e., model $\hat{\mM}$, in polynomial-time. Via such a solution, we obtain $\hat{p}_-\in \mathbb{R}^{\mS\mA}$ and can recover an estimate of the latent model $p_1(s,a) = \hat{R}_1(r=1|s,a)$ and $p_2(s,a) = \hat{R}_2(r=1|s,a)$ for all state-action pairs by a plug-in estimate~\eqref{eq: recovery of reward}. We refer to such a model as the {\em empirical model}. The proof of Lemma \ref{lemma:feasible_optimism} is given in Appendix \ref{appendix:optimism_feasibility}.

\begin{algorithm}[t]
    \caption{Reward Model Recovery from Second-Order Correlations}
    \label{algo:LMDP_recovery}
    \begin{algorithmic}[1]
        \STATE Solve an LP for $\hat{l}(x)$ and find $sign(x)$ for all $x \in \mS \times \mA$ that satisfies \eqref{eq:lp_variables}. 
        \STATE Clip $\hat{l} (x)$ within $(-\infty, \hat{u} (x)]$ where $\hat{u} (x) = \log \left(\min(\hat{p}_+(x), 1 - \hat{p}_+(x) \right)$ for all $x \in \mS \times \mA$.
        \STATE Set $\hat{p}_-(x) = sign(x) \cdot \exp(\hat{l}(x))$ for all $x \in \mS \times \mA$.
        \STATE Let $\hat{p}_1(x) = \hat{p}_+(x) + \hat{p}_-(x), \hat{p}_2(x) = \hat{p}_+(x) - \hat{p}_-(x)$ for all $x \in \mS \times \mA$.
        \STATE Return $\hat{\mM} = (\mS, \mA, \hat{T}, \hat{\nu}, \{\hat{R}_m\}_{m=1}^2)$, an empirical 2RM-MDP model.
    \end{algorithmic}
\end{algorithm}

\subsection{Main Theoretical Results}
\label{subsec:mainresults}
\paragraph{Upper Bound} We first prove the correctness of our main algorithm (Algorithm \ref{algo:two_reward_mixture}) and compute an upper bound on its sample complexity. 
Our upper bound has a quadratic dependence on $S \cdot A$. This is well-expected because we must estimate quantities related to the augmented second order MDP. Our lower bound shows that this dependence is in fact tight. Additionally, our bounds exhibit an interesting dependence on the error threshold, $\epsilon$, that in fact depends on the minimum separation in distinguishable state-actions:
\begin{align}
    \label{def:minimum_separation}
    \delta = 1 \wedge \min_{x \in \mS \times \mA: \Delta(x) > 0} \Delta(x).
\end{align}
where $\Delta(x) := |p_-(x)|$. This is reminiscent of similar results for mixture models (e.g., \cite{chen2017convex,kwon2020minimax}). 
We state the following theorem on the sample-complexity of 2RM-MDPs. 
\begin{theorem}
    \label{theorem:upper_bound}
    There exists a universal constant $C > 0$ such that if we run Algorithm \ref{algo:two_reward_mixture} with $\epsilon_{pe} = \frac{\epsilon}{H^3} \cdot \left(\delta \vee \frac{\epsilon}{H^2} \right)$, then Algorithm \ref{algo:pure_explore} terminates after at most $K$ episodes where,
    \begin{align*}
        K = C \cdot H^{6} \cdot \frac{S^2 A}{\epsilon^2} \cdot (H+A) \cdot \left(\frac{\epsilon}{H^2} \vee \delta\right)^{-2} \cdot \log(HSA/\epsilon \eta) \cdot \log^2(H/\epsilon),
    \end{align*}
    with probability at least $1 - \eta$. Furthermore, under the same high probabilistic event, Algorithm \ref{algo:two_reward_mixture} returns an $\epsilon$-optimal policy. 
\end{theorem}
Theorem \ref{theorem:upper_bound} shows that we can find a $\epsilon$-optimal policy for any 2RM-MDP instance with at most $O \left(poly(H, \epsilon^{-1}) \cdot S^2 A^2 \right)$ episodes of exploration. We leave it as future work to investigate whether dependency on $\epsilon$ can be uniformly $O(1/\epsilon^2)$ regardless of $\delta$, as well as whether the dependency on $H$ can be tightened.

\paragraph{Lower Bound} In Theorem \ref{theorem:upper_bound}, the sample complexity guaranteed by Algorithm \ref{algo:two_reward_mixture} is $\Omega(S^2 A^2)$. We show that the dependency on $S^2 A^2$ is also necessary for two reward-mixing MDPs:
\begin{theorem}
    \label{theorem:lower_bound}
    There exists a class of 2RM-MDPs such that in order to obtain $\epsilon$-optimal policy, we need at least $\Omega(S^2 A^2 / \epsilon^2)$ episodes. 
\end{theorem}
The proof of the lower bound can be found in Appendix \ref{appendix:lower_bound}. 

\paragraph{Computational Complexity} Main bottlenecks of Algorithm \ref{algo:two_reward_mixture} are in two parts: (1) solving the LP in Algorithm \ref{algo:LMDP_recovery} and (2) computing the optimal policy for $\hat{\mM}$. For (1), with $SA$ variables and $O(S^2 A^2)$ constraints, solving the LP in Algorithm \ref{algo:LMDP_recovery} takes $poly(S, A)$ times (see \cite{cohen2021solving} and the references therein for some known computational complexity results for solving LPs). For (2), point-based value-iteration algorithm \cite{pineau2006anytime} runs in $O(\epsilon^{-2} H^5 SA)$ time to obtain $\epsilon$-optimal policy for any specified 2RM-MDP model. Therefore, the overall running time of Algorithm \ref{algo:two_reward_mixture} is polynomial in all parameters.

%% file: Analysis.tex
For the ease of presentation, we assume that the true transition and initial state probabilities are known in advance. In Appendix \ref{Appendix:complete_upper_bound}, we complete the proof of Theorem \ref{algo:two_reward_mixture} without assuming known transition and initial probabilities but instead using $\hat{\nu}$ and $\hat{T}$ estimated during the pure-exploration phase.

We first note that given any policy $\pi \in \Pi$, difference in expected rewards can be bounded by $l_1$-statistical distance in trajectory distributions. More specifically, consider the set of all possible trajectories $\mT = (\mS \times \mA \times \{0,1\})^{\bigotimes H}$, that is, any state-action-reward sequence of length $H$. Then, 
$$|V_{\mM}^{\pi} - V_{\hat{\mM}}^{\pi}| \le H \cdot \|(\PP^\pi - \hat{\PP}^{\pi}) ((x,r)_{1:H})\|_1 = \sum_{\tau \in \mT} |\PP^\pi(\tau) - \hat{\PP}^\pi (\tau)|,$$
where $\hat{\PP}^{\pi} (\cdot)$ is the probability of an event measured in the empirical model $\hat{\mM}$ with policy $\pi$. Therefore, for any policy $\pi \in \Pi$, our goal is to show that ${\sum_{\tau \in \mT} |\PP^\pi (\tau) - \hat{\PP}^{\pi}(\tau)| \le O(\epsilon / H)}$, {\it i.e.,} the true and empirical models are close in $l_1$-statistical distance in trajectory distributions.


The main challenge in the analysis is to bound the $l_1$ distance when estimated first and second-order moments are within the range of confidence intervals. We now show that when all pairs of state-actions in $\tau$ were visited sufficient amount of times then we can bound the $l_1$ difference. Let us first divide a set of trajectories into several groups depending on the number of times that every pair in each trajectory has been explored during the pure-exploration phase. We first define the following sets: 
\begin{align}
    \label{eq:define_countlevel_sets}
    \mX_l &= \{ (x_i, x_j) \in (\mS\times \mA)^{\bigotimes 2} \ | \ n(x_i, x_j) \ge n_l\}, \nonumber \\
    \Eps_l &= \{ x_{1:H} \in \mT \ | \ \forall t_1, t_2 \in [H], \ t_1 \neq t_2 \ \text{ s.t. } \ (x_{t_1}, x_{t_2}) \in \mX_{l} \},
\end{align}
where $n_l = C \cdot \iota_2 \delta_l^{-2} \epsilon_l^{-2}$ for some absolute constant $C > 0$, $\epsilon_0 = \epsilon / H^2$ and $\epsilon_{l} = 2 \epsilon_{l-1}$ for $l \ge 1$. $\delta_l = \max(\delta, \epsilon_l)$ is a threshold for {\it distinguishable} state-actions $x$ such that $\Delta(x) \ge \delta_l$. In a nut-shell, $\Eps_l$ is a set of sequences of state-actions in which every pair of state-actions has been explored at least $n_l$ times. Then we split a set of trajectories into disjoint sets $\Eps_0' = \Eps_0$ and $\Eps_l' = \Eps_{l-1}^c \cap \Eps_l$ and for $l \in [L+1]$, {\it i.e.,} a set of trajectories with all pairs visited more than $n_l$ and at least one pair visited less than $n_{l-1}$, where we let $L = O(\log(H/\epsilon))$ be the largest integer such that $n_L \ge C \cdot \iota_2$. Let $n_{L+1} = 0$ and $\epsilon_{L+1} = 1$.

Recall that our goal is to control the $l_1$-statistical distance between distributions of trajectories for any history-dependent policies that resulted from true and empirical models. With above machinery, we can, instead, bound the $l_1$ statistical distance between all trajectories as the following:
\begin{align}
    \label{eq:bound_total_variation_distance_overall}
    \|(\PP^\pi - \hat{\PP}^\pi) (\tau)\|_1 = \sum_{l=0}^{L+1} \sum_{\tau:x_{1:H} \in \Eps_l'} |\PP^{\pi}(\tau) - \hat{\PP}^{\pi}(\tau) | \le \sum_{l = 0}^{L+1} \sup_{\pi \in \Pi} \PP^\pi( x_{1:H} \in \Eps_l' ) \cdot O(H \epsilon_l),
\end{align}
where $\PP^\pi( x_{1:H} \in \Eps_l'),$ is the probability that the random trajectory $\tau$ observed when following a policy $\pi$ is contained within the set $\Eps_{l}'$. The following lemma is critical in bounding the above inner sum for each fixed $l$: 
\begin{lemma}[Eventwise Total Variance Discrepancy]
    \label{lemma:eventwise_total_bound}
    For any $l \in \{0, 1, ..., L+1 \}$ and any history-dependent policy $\pi \in \Pi$, we have:
    \begin{align}
        \label{eq:eventwise_total_variance_diff}
        \sum_{\tau: x_{1:H} \in \Eps_l'} | \PP^\pi (\tau) - \hat{\PP}^\pi(\tau) | \le \sup_{\pi \in \Pi} \PP^\pi (x_{1:H} \in \Eps_l') \cdot O(H \epsilon_l). 
    \end{align}
\end{lemma}

The second key connection is the relation between $\sup_{\pi \in \Pi} \PP^{\pi} (\Eps_l')$ and the data collected in pure-exploration phase. The following lemma connects them:
\begin{lemma}[Event Probability Bound by Pure Exploration]
    \label{lemma:reward_free_guarantee}
    With probability at least $1-\eta$ Algorithm \ref{algo:pure_explore} terminates after at most $K$ episodes where
    \begin{align}
        \label{eq:required_episodes}
        K \ge C \cdot S^2A (H+A) \epsilon_{pe}^{-2} \log(HSA/(\epsilon_{pe} \eta)),
    \end{align}
    with some absolute constant $C > 0$. Furthermore, for every $l \in [L+1]$ we have 
    \begin{align}
        \sup_{\pi \in \Pi} \PP^\pi(x_{1:H} \in \Eps_{l}') \le O \left(H \epsilon_{pe} \delta_l^{-1} \epsilon_l^{-1} \right). \label{eq:max_event_probability}
    \end{align}
\end{lemma}
Given the above two lemmas, we set $\epsilon_{pe} = \epsilon \delta_0 / H^3 L$. We then apply Lemma \ref{lemma:reward_free_guarantee} and \eqref{eq:bound_total_variation_distance_overall} to bound a difference in expected value of true and empirical models:
\begin{align*}
    |V_\mM^\pi - V_{\hat{\mM}}^\pi| \le H \cdot \|(\PP^\pi - \hat{\PP}^\pi) (\tau)\|_1 \le \tilde{O} \left(H^3 \epsilon_{pe} \delta_0^{-1}\right) \le O(\epsilon). 
\end{align*}
Finally, plugging $\epsilon_{pe} = \epsilon \delta_0 / H^3 L$ into the required number of episodes \eqref{eq:required_episodes} gives a sample-complexity bound presented in Theorem \ref{theorem:upper_bound}. Full proof of Lemma \ref{lemma:reward_free_guarantee} is given in Appendix \ref{appendix:reward_free_guarantee}.

\subsection{Key Ideas for Lemma \ref{lemma:eventwise_total_bound}}
\label{subsec:lemma_41_overview}
The core of analysis by which we establish Lemma~\ref{lemma:eventwise_total_bound} goes as follows. We split the state-action pairs into two sets {\it distinguishable} and {\it indistinguishable} state-action pairs. Fix an $l \in \{0, 1, ... L+1\}$ and consider then set $\Eps_l'$. We call a state-action pair $x \in \mS \times \mA$ $\delta$-distinguishable if $\Delta(x)$ is larger then some $\delta > 0$. Then, by the definition of $\Delta(x)$, it implies that $p_1(x)$ and $p_2(x)$ are sufficiently different. For a trajectory $\tau \in \Eps_l'$, we say that a state-action $x$ within $\tau$ is $\delta_l$-distinguishable if $\Delta(x) \ge \delta_l$.

States that are distinguishable have the following property. If pairwise correlations between three distinguishable state-actions are well estimated, then we can recover individual reward probability for each of the three state-actions as in the following lemma:
\begin{lemma}
    \label{lemma:eps3_parameter_error}
    Suppose that $(x_1, x_2), (x_2, x_3), (x_3, x_1) \in \mX_l$ and $\Delta(x_i) \ge \delta_l$ for all $i \in \{1,2,3\}$. Let $i^* = \arg \max_{i\in [3]} \Delta(x_i)$, and $i_2^* = arg \max_{i \in [3], i\neq i^*} \Delta(x_i)$. Then we have
    \begin{align*}
        \forall i \neq i^*, \ | \hat{\Delta} (x_i) - \Delta(x_i) | \le \epsilon_l / 2, \textit{ and }
        \left| \hat{\Delta} (x_{i^*}) - \Delta(x_{i^*}) \right| \le \frac{\Delta_{i^*}}{\Delta_{i_2^*}} \epsilon_l / 2.
    \end{align*}
\end{lemma}
If all reward parameters in a trajectory are relatively well-estimated, then we can bound the error in the estimated and true probability of the trajectory. Thus, if a trajectory $\tau \in \Eps_l'$ contains more than two distinguishable state-actions, we can rely on the closeness in reward parameters to bound $|\PP^\pi (\tau) - \hat{\PP}^{\pi} (\tau)|$. On the other hand, if there are no such three distinguishable state-actions in $\tau$, then we cannot guarantee the correctness of reward parameters in $\tau$. In such case, we take a different path to show the closeness in trajectory distributions. Full proof is given in Appendix \ref{appendix:eventwise_total_bound}.

%% file: Appendix.tex
\section{Proof of Lemma \ref{lemma:eventwise_total_bound}}
\label{appendix:eventwise_total_bound}

\subsection{Notation}
We define new notation that will be utilized across the appendix. Let $\Eps$ be a subset of trajectories such that for some mapping $f: (\mS \times \mA)^{\bigotimes H} \rightarrow \{0,1\}$, let $\Eps = \{x_{1:H} | f(x_{1:H}) = 1\}$. Let us define $\PP_\Eps^* (h_t)$ be the maximum probability of getting any final trajectory $\tau = (x,r)_{1:H}$ that belongs to $\Eps$ starting from a history $h_t = ((x,r)_{1:t-1}, s_t)$. That is, 
\begin{align}
    \label{eq:def_eps_*}
    \PP_\Eps^* (h_t) = \sup_{\pi \in \Pi} \PP^\pi (x_{1:H} \in \Eps | h_t). 
\end{align}
At the beginning of the episode, $h_0 = \phi$ and we denote $\PP_{\Eps}^* (\phi)$ to mean $\sup_{\pi \in \Pi} \PP^{\pi} (x_{1:H} \in \Eps)$. We often use $\PP^{\pi} (\Eps)$ instead of $\PP^{\pi} (x_{1:H} \in \Eps)$, omitting $x_{1:H}$ when the context is clear.

We use $\pi_{1:t}$ for any $t \in [H]$ to denote $\Pi_{t'=1}^t \pi(a_{t'} | h_{t'})$. Similarly, with a slight abuse in notation, we define $T_{0:t} := \nu(s_1) \cdot \Pi_{t'=1}^{t} T(s_{t'+1} | s_{t'}, a_{t'})$ for $1 \le t \le H-1$. When we define a new RM-MDP model $\mM^c$ with a superscript $c \in \mathbb{N}$, we add a superscript $c$ for all quantities measured with respect to $\mM^c$. For instance, reward probability at state $s$ and action $a$ in the $m^{th}$ in a model $\mM^a$ is denoted as $R_m^c (\cdot | s,a)$, a probability of an event in a model $\mM^a$ with a policy $\pi$ is denoted as $\PP^{c, \pi} (\cdot)$. 

In order to express the probability of a trajectory without conditioning on the context, we first introduce a convenient matrix form of probability $D(r|x) = \begin{bmatrix} R_1(r|x) & 0 \\ 0 & R_2(r|x) \end{bmatrix}$. Probability of observing a trajectory $\tau = (x,r)_{1:H}$ is then 
$$\PP^\pi (\tau) = \frac{1}{2} \pi_{1:t} T_{0:t} \cdot \mathbf{1}^\top \left(\Pi_{t=1}^H D(r_t|x_t)\right) \mathbf{1},$$ 
where $\mathbf{1}$ is the all-one vector $[1 1]^\top$. For any $t_1, t_2 \in [H]$, let $D_{t_1:t_2}$ be a short hand for $\Pi_{t=t_1}^{t_2} D(r_{t} | x_{t})$. We use $(\cdot)_{t_1:t_2}$ in a similar manner for any other symbols. If $t = t_1 = t_2$, then we simply use $(\cdot)_{t}$. For instance, we use $D_t = D(r_t|x_t) = \begin{bmatrix} R_1(r_t|x_t) & 0 \\ 0 & R_2(r_t|x_t) \end{bmatrix}$.

\subsection{Proof of Lemma \ref{lemma:eventwise_total_bound}}
We provide the outline of the proof in this section. All omitted proofs can be found in Appendix~\ref{appendix:auxiliary_lemmas}. To simplify the presentation, we temporarily assume that we know the transition and initial state probabilities, {\it i.e.,} $\hat{T} = T, \hat{\nu} = \nu$. In Appendix \ref{appendix:subsec:complete_upper_bound}, we provide the full proof without assuming known transition and initial state probabilities, but intead using $\hat{\nu}$ and $\hat{T}$ estimated in Algorithm \ref{algo:pure_explore}. 

The analysis begins with the following lemma on the summation of target quantity when differences in two models are small parameter-wise. 
\begin{lemma}
    \label{lemma:basic_tot_bound}
    Suppose that two 2RM-MDPs $\mM^1, \mM^2$ have the same transition kernel and initial distribution, and satisfy $\|(R_m^1 - R_m^2) (r | x) \|_1 \le \epsilon_r$ for some $\epsilon_r > 0$ and any $m \in \{1,2\}$ and $x \in \mX$ for a given set $\mX \subseteq \mS \times \mA$. For any subset of length $H$ state-action sequences $\Eps \subseteq \mX^{\bigotimes H}$, we have
    \begin{align*}
        \sup_{\pi \in \Pi} \sum_{\tau: x_{1:H} \in \Eps} | \PP^{1,\pi}(\tau) - \PP^{2,\pi} (\tau) | &\le \sup_{\pi \in \Pi} \PP^{1,\pi} (\Eps) \cdot H \epsilon_r,
    \end{align*}
    where $\PP^{c, \pi} (\cdot)$ is a probability measured in an environment modeled by $\mM^c$ for $c = 1,2$. 
\end{lemma}
Given Lemma \ref{lemma:basic_tot_bound}, we first consider surrogate 2RM-MDPs for the true and estimated models which ignore small differences in reward functions. Specifically, for all $x \in \mS \times \mA$, we define $\mM^1$ that approximates the true model as: 
\begin{align}
    \mM^1: & p_m^1 (x) = \epsilon_l + (1 - 2\epsilon_l) p_+ (x),  \quad  m \in \{1,2\}, &\text{if $\Delta (x) < 2\epsilon_l$}, \nonumber \\
    & p_m^1 (x) = \epsilon_l + (1 - 2\epsilon_l) p_m(x),  \quad  m \in \{1,2\},  & \text{if $\Delta (x) \ge 2\epsilon_l$}, \label{eq:define_surrogate_m1}
\end{align}
and similarly $\mM^2$ for the estimated model: 
\begin{align}
    \mM^2: & p_m^2 (x) = \epsilon_l + (1 - 2\epsilon_l) p_+ (x), \quad  m \in \{1,2\}, &\text{if $ \hat{\Delta} (x) < 2\epsilon_l$}, \nonumber  \\
    & \begin{cases} p_1^2 (x) = \epsilon_l + (1 - 2\epsilon_l) (p_+ + \hat{p}_-)(x) \\ p_2^2 (x) = \epsilon_l + (1 - 2\epsilon_l) (p_+ - \hat{p}_-)(x) \end{cases}, & \text{if $ \hat{\Delta} (x) \ge 2\epsilon_l$}. \label{eq:define_surrogate_m2}
\end{align}
Recall that reward models are determined by $R_m^c (r = 1|x) = p_m^c(x)$ and $R_m^c(r=0 | x) = 1 - p_m^c(x)$ for $m = 1,2$ and $c = 1,2$. In the above construction, note that for the average value we use the same quantity $p_+$ from the true model to ignore small differences in averaged rewards. In particular, note that $p_m^1(x) = p_m^2(x)$ for any $x$ with $\Delta(x) < 2\epsilon_l$ and $\hat{\Delta}(x) < 2\epsilon_l$. For both models, we use true transition and initial state distribution models $T$ and $\nu$. By construction $\mM^1$ and $\mM^2$ well approximate $\mM^*$ and $\hat{\mM}$ respectively:
\begin{lemma}
    \label{lemma:surrogate_models_proximity}
    For all $m \in \{1,2\}$ and $\forall x: \exists (x_i, x_j) \in \mX_l \text{ s.t. } x = x_i \text{ or } x_j$, we have 
    \begin{align*}
        \| (R_m - R_m^1) (r|x) \|_1 \le 4\epsilon_l, &\quad \| (\hat{R}_m - R_m^2) (r|x) \|_1 \le 4\epsilon_l.
    \end{align*} 
\end{lemma}
Then using Lemma \ref{lemma:basic_tot_bound}, we have
\begin{align*}
    \sum_{\tau:x_{1:H} \in \Eps_l'} | \PP^\pi (\tau) - \hat{\PP}^\pi(\tau) | &\le  \sup_{\pi \in \Pi} \PP^\pi (\Eps_l') \cdot O(H \epsilon_l) + \sum_{\tau:x_{1:H} \in \Eps_l'} | \PP^{1,\pi} (\tau) - \PP^{2,\pi}(\tau) |, 
\end{align*}
where we recall that $\Eps_l'$ is defined as $\Eps_0' = \Eps_0,$ and $\Eps_l' = \Eps_{l-1}^c \cap \Eps_l$ for $l \ge 1$.

Now we continue the discussion in Section \ref{subsec:lemma_41_overview}. We divide a set of trajectories $\Eps_l'$ by whether the number of $\delta_l$-distinguishable pairs in a trajectory $\tau: x_{1:H} \in \Eps_l'$ is less than 3, {\it i.e.,} $\sum_{t=1}^H \indic {\Delta(x_t) \ge \delta_l}$ being $\le 2$ or $\ge 3$:
\begin{align}
    \Epsd &= \left\{x_{1:H} \in \Eps_l' \Big| \sum_{t=1}^H \indic{\Delta(x_t) \ge \delta_l} \le 2 \right\}, \nonumber \\  
    \Epst &= \left\{x_{1:H} \in \Eps_l' \Big| \sum_{t=1}^H \indic{\Delta(x_t) \ge \delta_l} \ge 3 \right\}. \label{eq:define_eps_23}
\end{align}
We handle each case separately and show that $\sum_{\tau: x_{1:H} \in \Eps} |\PP^{1,\pi} (\tau) - \PP^{2,\pi} (\tau)| \le \sup_{\pi \in \Pi} \PP^\pi(\Eps) \cdot  O(H\epsilon_l)$ for $\Eps = \Epsd$ and $\Epst$ respectively.

\subsubsection{Case I: $\Epst$.} 
\label{subsubsection:proof_overview_eps3}
For any $\tau: x_{1:H} \in \Epst$, let any $t_1, t_2, t_3 \in [H]$ such that $\Delta(x_{t_i}) \ge \delta_l$ for $i=1,2,3$. From Lemma~\ref{lemma:eps3_parameter_error}, we obtain a corollary on parameters of all state-actions that appear in $\tau: x_{1:H} \in \Epst$:
\begin{corollary}
    \label{corollary:big_delta_bound}
    For any $\tau: x_{1:H} \in \Epst$ and for every $t \in [H]$, let $t^* = \arg \max_{t \in [H]} \Delta(x_t)$, and $t_2^* = arg \max_{t \in [H], t \neq t^*} \Delta(x_t)$. Then we have
    \begin{align*}
        |\hat{\Delta}(x_t) - \Delta(x_t)| \le 3\epsilon_l \quad \forall t \neq t^*, \qquad
        \left| \hat{\Delta}(x_{t^*}) - \Delta(x_{t^*}) \right| \le \frac{\Delta_{t^*}}{\Delta_{t_2^*}} \epsilon_l/2.
    \end{align*}
\end{corollary}
That is, in all trajectories in $\Epst$, all visited state-actions have good estimates of reward probabilities within $O(\epsilon_l)$-error (except at most one state-action $x_{t^*}$, which needs an extra care). In particular, we work with the fact that $|\Delta^1(x_t) - \Delta^2(x_t)| < O(\epsilon_l)$ for most of $t \in [H]$. This case is handled in Appendix \ref{appendix:complete_eps3}.

\subsubsection{Case II: $\Epsd$.} 
\label{subsubsection:proof_overview_eps2}
In this case, the following lemma is the key result to bound the error in this case. 
\begin{lemma}
    \label{lemma:eps2_count_distinguishable}
    For any $x_{1:H} \in \Epsd$, we have
    $\sum_{t=1}^H \indic{ \Delta(x_t) \ge \epsilon_l \cup \hat{\Delta} (x_t) \ge 2\epsilon_l } \le 2$.
\end{lemma}
Lemma \ref{lemma:eps2_count_distinguishable} ensures that for any $x_{1:H} \in \Epsd$, except at most for some two time steps $t_1 ,t_2 \in [H]$, we have $\Delta(x_t), \hat{\Delta}(x_t)$ indistinguishable for all $t \neq t_1, t_2$. For such $t \neq t_1, t_2$, since an average of rewards can be well-estimated, {\it i.e.,} $p_+(x_t) \approx \hat{p}_+ (x_t)$, let us for now ignore the difference at $x_t$: $R_m(\cdot | x_t) \approx \hat{R}_m(\cdot | x_t)$ for $m = 1,2$. Then for any $\tau = (x,r)_{1:H}$ for which $x_{1:H}$ belongs to $\Epsd$:
\begin{align*}
    |\PP^\pi \left(\tau \right) - \hat{\PP}^\pi (\tau)| \propto \PP^\pi(\tau) \cdot |\mu(x_{t_1}, x_{t_2}) - \hat{\mu}(x_{t_1}, x_{t_2})| \le O(\epsilon_l),
\end{align*}
where $\mu(x_{t_1},x_{t_2})$ is from equation \eqref{eq:def_mu}, and the last inequality is due to the construction of $\hat{\mM}$ which is designed to match in second-order correlations. Building upon the above idea, we show that $\sum_{\tau: x_{1:H} \in \Epsd} |\PP^{\pi} (\tau) - \hat{\PP}^{\pi} (\tau) | \le O(H \epsilon_l) \cdot \sup_{\pi \in \Pi} \PP^\pi (\Epsd)$. This case is handled in Appendix \ref{appendix:complete_eps2}. Once we prove for $\Epsd$ and $\Epst$ such that
\begin{align}
    \sum_{\tau: x_{1:H} \in \Epst} |\PP^{1,\pi} (\tau) - \PP^{2,\pi} (\tau)| \le O(H \epsilon_l) \cdot \sup_{\pi \in \Pi} \PP^{\pi} (\Epst), \label{eq:eps_3_bound} \\ 
    \sum_{\tau: x_{1:H} \in \Epsd} |\PP^{1,\pi} (\tau) - \PP^{2,\pi} (\tau)| \le O(H \epsilon_l) \cdot \sup_{\pi \in \Pi} \PP^{\pi} (\Epsd),
    \label{eq:eps_2_bound}
\end{align}
then, since $\Epsd \cup \Epst = \Eps_l'$, we obtain $\sum_{\tau: x_{1:H} \in \Eps_l'} | \PP^\pi (\tau) - \hat{\PP}^\pi(\tau) | \le O(H \epsilon_l) \cdot \sup_{\pi \in \Pi} \PP^{\pi} (\Eps_l')$, which concludes Lemma \ref{lemma:eventwise_total_bound}. 

\subsection{Proof of Lemma \ref{lemma:basic_tot_bound}}
\label{appendix:lemma:basic_tot_bound}

By the definition of $\PP_{\Eps}^*(\cdot)$ defined in \eqref{eq:def_eps_*}, we have the following inequalities: for any length $t$ history $h_t = ((s,a,r)_{1:t-1}, s_t)$, action $a_t$ and any history-dependent policy $\pi$, we have
\begin{align}
    \PP_\Eps^* (h_{t}) \ge \sum_{a_t} \PP_\Eps^*(h_{t}, a_t) \pi(a_t| h_t), &\ \quad t < H, \label{lemma:bellman_for_max_event_probability1} \\
    \PP_\Eps^* (h_{t}, a_t) \ge \sum_{s_{t+1}} \PP_\Eps^*(h_{t+1}) T(s_{t+1} | s_t, a_t), &\ \quad t < H, \label{lemma:bellman_for_max_event_probability2} \\
    \PP_\Eps^* (h_{H}) \ge \sum_{a_H} \indic{x_{1:H} \in \Eps} \pi(a_H| h_H), &\ \quad t = H. \label{lemma:bellman_for_max_event_probability3}
\end{align}
Also, since $\PP_{\Eps^*} = \sup_{\pi} \PP^{\pi} (\tau \in \Eps)$ only depends on the occurance of $x_{1:H}$, any two 2RM-MDP models with the same transition and initial distribution have the same value for $\PP_{\Eps^*}$:
\begin{align*}
    \PP_\Eps^* (h) = \sup_{\pi \in \Pi} \PP^{1,\pi} (x_{1:H} \in \Eps|h) = \sup_{\pi \in \Pi} \PP^{2,\pi} (x_{1:H} \in \Eps|h).
\end{align*}
Hence when we consider the same transition model, we often omit $1$ and $2$ in superscript from $\PP^{1,\pi}(\Eps| h)$ or $\PP^{2,\pi} (\Eps| h)$.

\begin{proof}
    Now we prove Lemma \ref{lemma:basic_tot_bound}. Our target is to analyze the difference in the following: $$\PP^{c,\pi}(\tau) = \nu(s_1) \left(\Pi_{t=1}^H \pi(a_t|h_t) \right) \cdot \left(\Pi_{t=1}^{H-1} T(s_{t+1} | s_t,a_t)\right) \cdot \frac{1}{2} \mathbf{1}^\top D^c_{H:1} \mathbf{1},$$ for $c = 1,2$ for all $\tau: x_{1:H} \in \Eps$. 
    
    We prove the lemma by backward induction. Let us denote $h_t = ((s,a,r)_{1:t-1}, s_{t})$. Note that we assume here for any $x_t = (s_t,a_t)$ that appear in a trajectory $x_{1:H} \in \Eps$ at any time $t \in [H]$ satisfies $\| D^1(r_t|x_t) - D^2(r_t|x_t) \|_1 < \epsilon_r$. Hence,
    \begin{align*}
        &2 \cdot \sum_{\tau: x_{1:H} \in \Eps} | \PP^{1,\pi}(\tau) - \PP^{2,\pi}(\tau) | \\
        &= \sum_{h_H} \sum_{a_H, r_H} \indic{x_{1:H} \in \Eps} \pi(a_H | h_H) \pi_{1:H-1} T_{0:H-1} | \mathbf{1}^\top (D^1_{H:1} - D^2_{H:1}) \mathbf{1} | \\
        &\le \sum_{h_H} \sum_{a_H,r_H} \indic{x_{1:H} \in \Eps} \pi(a_H | h_H) \pi_{1:H-1} T_{0:H-1} \|(D_H^1 - D_H^2) \cdot D_{H-1:1}^1 \mathbf{1}\|_1 \\
        &\ + \sum_{h_H} \sum_{a_H} \indic{x_{1:H} \in \Eps} \pi(a_H | h_H) \pi_{1:H-1} T_{0:H-1} \|D_{H}^2 (D_{H-1:1}^1 - D_{H-1:1}^2) \mathbf{1} \|_1 \\
        &\le \sum_{h_H} \sum_{a_H} \indic{x_{1:H} \in \Eps} \pi(a_H | h_H) \pi_{1:H-1} T_{0:H-1} \|D^1_{H-1:1} \mathbf{1}\|_1 \epsilon_r \\
        &\ + \sum_{h_H} \sum_{a_H} \indic{x_{1:H} \in \Eps} \pi(a_H | h_H) \pi_{1:H-1} T_{0:H-1} \|(D_{H-1:1}^1 - D_{H-1:1}^2) \mathbf{1} \|_1 \\
        &\le \epsilon_r \sum_{h_H} \PP_{\Eps}^*(h_H) \PP^{1,\pi}(h_H) + \sum_{h_H} \PP_\Eps^*(h_H) \pi_{1:H-1} T_{0:H-1} \|(D_{H-1:1}^1 - D_{H-1:1}^2) \mathbf{1} \|_1.
    \end{align*}
    The first term in the last inequality above can be bounded as the summation over all possible length $H$ histories such that for any policy $\pi \in \Pi$, we have
    \begin{align*}
        \PP^*_\Eps (\phi) = \sup_{\pi} \PP^{\pi} (\Eps) = \sup_{\pi} \sum_{h_H} \PP^\pi (\Eps,h_H) = \sup_{\pi} \sum_{h_H} \PP^\pi (\Eps|h_H) \PP^\pi(h_H) = \sup_{\pi} \sum_{h_H} \PP_\Eps^*(h_H) \PP^\pi(h_H).
    \end{align*}
    Then we can proceed from time step $t = H$ to $t = H-1$:
    \begin{align*}
        &\sum_{h_H} \PP_\Eps^*(h_H) \pi_{1:H-1} T_{0:H-1} \|(D_{H-1:1}^1 - D_{H-1:1}^2) \mathbf{1} \|_1 \\
        &\le \sum_{h_{H-1}} \sum_{a_{H-1}, r_{H-1}, s_H} \PP_\Eps^*(h_H) \pi(a_{H-1} | h_{H-1}) T(s_H | x_{H-1}) \pi_{1:H-2} T_{0:H-2} \|(D_{H-1}^1 - D_{H-1}^2) \cdot D_{H-2:1}^1 \mathbf{1} \|_1 \\
        &+ \sum_{h_{H-1}} \sum_{a_{H-1}, s_H} \PP_\Eps^*(h_H) \pi(a_{H-1} | h_{H-1}) T(s_H | x_{H-1}) \pi_{1:H-2} T_{0:H-2} \|(D_{H-2:1}^1 - D_{H-2:1}^2) \mathbf{1}\|_1 \\ 
        &\le \epsilon_r \cdot \sum_{h_{H-1}} \PP^{1,\pi} (h_{H-1}) \sum_{a_{H-1}, s_H} \PP_\Eps^*(h_H) \pi(a_{H-1} | h_{H-1}) T(s_H | x_{H-1})  \\
        &+ \sum_{h_{H-1}} \pi_{1:H-2} T_{0:H-2} \|(D_{H-2:1}^1 - D_{H-2:1}^2 ) \mathbf{1}\|_1 \sum_{a_{H-1},s_H} \PP_\Eps^*(h_H) \pi(a_{H-1} | h_{H-1}) T(s_H | x_{H-1}) \\
        &\le \PP_\Eps^*(\phi) \epsilon_r + \sum_{h_{H-1}} \PP_\Eps^* (h_{H-1}) \pi_{1:H-2} T_{0:H-2} \|(D_{H-2:1}^1 - D_{H-2:1}^2) \mathbf{1}\|_1,
    \end{align*}
    where the last inequality comes from \eqref{lemma:bellman_for_max_event_probability3}. We can repeat this procedure until we reach $t = 1$ in backwards. Note that $\PP_\Eps^* (\phi) = \sup_{\pi} \PP^\pi(\Eps)$, which gives Lemma \ref{lemma:basic_tot_bound}. 
\end{proof}

\subsection{Analysis for Case I: $\Epst$ (equation \eqref{eq:eps_3_bound})}
\label{appendix:complete_eps3}

\subsubsection{Equivalence in Signs} 
\label{appendix:equivalence_in_signs}
Before we start the proof, we need to point out one important fact. For any $x_{1:H} \in \Epst$, for all $x \in \{x_t\}_{t=1}^H$ such that $\Delta(x) \ge 2 \epsilon_l$, either one of the two is true: $sign(p_-(x)) = sign(\hat{p}_-(x))$ or $sign(p_-(x)) = -sign(\hat{p}_-(x))$. This then implies $sign(p_-^1(x)) = sign(p_-^2(x))$ or $sign(p_-^1(x)) = -sign(p_-^2(x))$ whenever $|p_-^1(x)| > 0$ and $|p_-^2(x)| > 0$ consistently for all state-actions $x$ of interest. This can be shown by a simple contradiction argument: suppose there exists $t_1, t_2 \in [H]$ such that $t_1 \neq t_2$ and $\Delta(x_{t_1}), \Delta(x_{t_2}) \ge 2\epsilon_l$. If $sign(\hat{p}_-(x_{t_1}) \hat{p}_-(x_{t_2})) \neq sign(p_-(x_{t_1}) p_-(x_{t_2}))$, then this implies
\begin{align*}
    | p_-(x_{t_1}) p_-(x_{t_2}) - \hat{p}_-(x_{t_1}) \hat{p}_-(x_{t_2}) | \ge \max(\delta^2, 4\epsilon_l^2) > \epsilon_l \delta_l.
\end{align*}
This violates the confidence interval constraint \eqref{eq:lp_variables} since $n(x_{t_1}, x_{t_2}) \ge n_l \ge C \cdot \iota_2 \delta_l^{-2} \epsilon_l^{-2}$ and thus $b(x_{t_1}, x_{t_2}) < 0.01 \delta_l \epsilon_l$, which forces $| p_-(x_{t_1}) p_-(x_{t_2}) - \hat{p}_-(x_{t_1}) \hat{p}_-(x_{t_2}) | < \delta_l \epsilon_l$.

Now if this is the case, then without loss of generality, we can assume $sign(p_-^1 (x)) = sign(p_-^2(x))$ for $x: |p_-^1(x)|, |p_-^2(x)| > 0$ since
\begin{align*}
    \PP^{2,\pi} (\tau) &= \frac{1}{2} \pi_{1:H} T_{0:H-1} \one^\top D_{H:1}^2 \one \\
    &= \frac{1}{2} \pi_{1:H} T_{0:H-1} \left(\Pi_{t=1}^H (p_+^2(x_t) + p_-^2(x_t)) + \Pi_{t=1}^H (p_+^2(x_t) - p_-^2(x_t)) \right),
\end{align*}
remains the same after we replace $p_-^2(x)$ by $-p_-^2(x)$ for all $x \in \{x_t\}_{t=1}^H$. This means, regardless of the sign of $p_-^2(x)$, the probability of any trajectories, which we eventually need, remains the same. Hence now, without loss of generality, we assume that $sign(p_{-}^1 (x)) = sign(p^2_-(x))$ for all $x \in \{x_t\}_{t=1}^H$ for any $x_{1:H} \in \Eps_{l,3}$. (Note that for non-uniform mixing weights, we need an extra care since $\PP^{2,\pi}(\tau)$ is no more symmetric in signs of $p^2_-(x)$. See Appendix \ref{appendix:sign_ambiguity} to see the discussion on handling non-uniform priors).

Once the above holds, we can claim that
\begin{align*}
    \|D^1(r=1|x) - D^2(r=1|x)\|_1 \le O(\epsilon_l),
\end{align*}
where $\|\cdot\|_1$ is a matrix $l_1$ norm here, for all $x \in \{x_t\}_{t=1}^H$. To see this, first note that 
\begin{align*}
    |R^1_1(r=1|x) - R^2_1(r=1|x)| &= |p_1^1(x) - p_1^2(x)| \\ 
    &= |(p_+^1 (x) + p_-^1 (x)) - (p_+^2(x) + p_-^2(x))| = |p^1_-(x) - p^2_-(x)|, 
\end{align*}
where the last equality comes from the construction \eqref{eq:define_surrogate_m1}, \eqref{eq:define_surrogate_m2} such that $p_+^1(x) = p_+^2(x)$. Since the sign of $p_-^1(x)$ is equal to the sign of $p_-^2(x)$, we have
\begin{align*}
    |p_-^1(x) - p_-^2(x)| = |\Delta^1(x) - \Delta^2(x)|.
\end{align*}
The same argument holds for $r = 0$. Note that (again from the construction), $|\Delta^1(x) - \Delta^2(x)| \le |\Delta(x) - \hat{\Delta}(x)| + 4\epsilon_l$ and thus whenever $\hat{\Delta}(x)$ is close to $\Delta(x)$, we have small error in $D^1(r|x)$ and $D^2(r|x)$ as well.

\subsubsection{Notation}
Before we proceed, we define a few more notation here. We occasionally use $R_+(r|x) = \frac{1}{2}(R_1^1 + R_2^1) (r|x)$, and $R_-^c (r|x) = \frac{1}{2} (R_1^c - R_2^c) (r|x)$ for $c = 1,2$. Let $D_+^M (r|x)$ and $D_-^M(r|x)$ for all $r \in \{0,1\}$ and $x \in \mS \times \mA$ be the model average and model difference defined as follows: 
\begin{align}
    D_+^M(r|x) &:= \frac{1}{2} (D^1 + D^2) (r|x) = \begin{bmatrix} (R_1^1 + R_1^2)(r|x) & 0 \\ 0 & (R_2^1 + R_2^2)(r|x) \end{bmatrix}, \\
    D_-^M(r|x) &:= \frac{1}{2} (D^1 - D^2)(r|x) = \frac{1}{2} (R_-^1 - R_-^2) (r | x) \begin{bmatrix} 1 & 0 \\ 0 & -1 \end{bmatrix},
\end{align} 
We also introduce $D_+(r|x) := R_+(r|x) \begin{bmatrix} 1 & 0 \\ 0 & 1\end{bmatrix}$.

We let $t^*, t_2^*$ as defined in Corollary \ref{corollary:big_delta_bound}: 
\begin{align*}
    t^* = \arg \max_{t \in [H]} \Delta(x_t), \qquad t^*_2 = \arg \max_{t \in [H] / \{t^*\}} \Delta(x_t).
\end{align*}
Let us consider a set of state-actions $\mXt$ such that for all $\tau \in \Epst$, they are always the only maximum distinguishable state-actions whenever they are included in a trajectory. Formally, we consider
\begin{align}
    \mXt = \{x | \forall x_{1:H} \in \Epst: \ x \notin \{x_t\}_{t=1}^H \text{ or } \exists t^* \in [H] \ s.t. \ x_{t^*} = x \text{ and } \Delta(x_t) \le \Delta(x) / 2, \ \forall t \neq t^* \}. 
\end{align}
Note that for all other state-actions that are not in $\mXt$ and appear in any trajectories in $\Epst$, the error between true $\Delta$ and estimated $\hat{\Delta}$ is less than $2\epsilon_l$ by Corollary \ref{corollary:big_delta_bound}. We define a set of trajectories 
\begin{align}
    \label{eq:eps3_def_eps4}
    \Epsq = \{x_{1:H} \in \Epst | \exists t^* \in [H], x_{t^*} \in \mXt \},
\end{align}
that contains one of state-actions in $\mXt$. 

\subsubsection{Main Proof} 
Now we focus on bounding the sum of errors in predictions of trajectories in $\Epst$. Our strategy is first to split trajectories into $\Eps_{l,4}$ and $\Eps_{l,3} \cap \Eps_{l,4}^c$:
\begin{align*}
    2 \cdot \sum_{\tau \in \Epst} | \PP^{1,\pi}(\tau) - \PP^{2,\pi}(\tau) | &= \sum_{(x,r)_{1:H}} \indic{x_{1:H} \in \Epst} T_{0:H-1} \pi_{1:H} | \mathbf{1}^\top (D^1_{H:1} - D^2_{H:1}) \mathbf{1} | \\
    &= \sum_{(x,r)_{1:H}} \indic{x_{1:H} \in \Epst \cap \Eps_{l,4}^c } T_{0:H-1} \pi_{1:H} | \mathbf{1}^\top (D^1_{H:1} - D^2_{H:1}) \mathbf{1} | \\
    &\ + \sum_{(x,r)_{1:H}} \indic{x_{1:H} \in \Eps_{l,4}} T_{0:H-1} \pi_{1:H} | \mathbf{1}^\top (D^1_{H:1} - D^2_{H:1}) \mathbf{1} |.
\end{align*}
For the first term, by Corollary \ref{corollary:big_delta_bound} for all $r\in\{0,1\}$ and $x \notin \mXt$ we have 
\begin{align*}
    \|D^1(r|x) - D^2(r|x)\|_1 &= \max_{m \in \{1,2\}} |R_m^1(r|x) - R_m^2(r|x)| \\
    &= |\Delta^1(x) - \Delta^2(x)| \le |\Delta(x) - \hat{\Delta}(x)| + 4\epsilon_l \le 8\epsilon_l.    
\end{align*} 
Now, since all trajectories in $\Epst \cap \Eps_{l,4}^c$ does not contain any state-actions in $\mXt$, we can use Lemma \ref{lemma:basic_tot_bound} to show that the first term is less than $\PP^*_{\Epst \cap \Eps_{l,4}^c}(\phi) \cdot O(H\epsilon_l)$. Therefore we focus on bounding the second term, the sum of errors in predictions of trajectories in $\Epsq$.

For the rest of the analysis, we aim to to get as tight sample-complexity bound as possible in terms of $\epsilon$. At a high level, errors from $x \in X_{l,3}$ can be potentially large when all other state-actions in trajectories have small gaps (Corollary \ref{corollary:big_delta_bound}). To handle this, we first introduce the following five new quantities: 
\begin{align*}
    &D^{1,0}(\cdot|x) = \begin{cases} D_+^M (\cdot|x), \quad \text{if } x \in \mXt \\ D^{1}(\cdot|x), \quad \text{otherwise} \end{cases}, \quad D^{2,0}(\cdot|x) = \begin{cases} D_+^M (\cdot|x), \quad \text{if } x \in \mXt \\ D^{2}(\cdot|x), \quad \text{otherwise} \end{cases}, \\
    &D^{1,1}(\cdot|x) = \begin{cases} D_-^M (\cdot|x), \quad \text{if } x \in \mXt \\ D^{1}(\cdot|x), \quad \text{otherwise} \end{cases}, \quad D^{2,1}(\cdot|x) = \begin{cases} D_-^M (\cdot|x), \quad \text{if } x \in \mXt \\ D^{2}(\cdot|x), \quad \text{otherwise} \end{cases}, \\
    &D^{3}(\cdot|x) = \begin{cases} D_-^M (\cdot|x), \quad \text{if } x \in \mXt \\ D_+ (\cdot|x), \quad \text{otherwise} \end{cases}.
\end{align*}
We first ignore the errors from $x \in X_{l,3}$ by bounding with $D^{1,0} - D^{2,0}$. Then, the remaining term becomes $D^{1,1} + D^{2,1}$, and this term pertains to estimation errors in $x \in X_{l,3}$. We then introduce an auxiliary variable so that we can bound this with $D^{1,1} - D^3$ and $D^{2,1} - D^3$. Note that $\bm{1}^\top D^3 \bm{1} = 0$ by the construction of $D^3$. The key idea is, if all other state-actions $x \notin X_{l,3}$ in each trajectory have small gaps, then $D^1 - D_+$ would be small  which can cancel out large estimation errors (contained in $D_-^M$) coming from $x \in X_{l,3}$.

Specifically, we decompose the target quantity into three:
\begin{align}
    \label{eq:eps3_three_decomposition}
    \sum_{x_{1:H} \in \Epsq, r_{1:H}} T_{0:H-1} \pi_{1:H} | \mathbf{1}^\top (D^1_{H:1} - D^2_{H:1}) &\mathbf{1} | \le \sum_{x_{1:H} \in \Epsq, r_{1:H}} T_{0:H-1} \pi_{1:H} | \mathbf{1}^\top (D^{1,0}_{H:1} - D^{2,0}_{H:1}) \mathbf{1} | \nonumber \\
    &+ \sum_{x_{1:H}\in \Epsq, r_{1:H}} T_{0:H-1} \pi_{1:H} | \mathbf{1}^\top (D^{1,1}_{H:1} - D^{3}_{H:1}) \mathbf{1} | \nonumber \\
    &+ \sum_{x_{1:H}\in \Epsq, r_{1:H}}  T_{0:H-1} \pi_{1:H} | \mathbf{1}^\top (D^{2,1}_{H:1} - D^{3}_{H:1}) \mathbf{1} |.
\end{align}
Since $\|D^{1,0}_t - D^{2,0}_t\|_1 \le O(\epsilon_l)$ for all $t$, the first term can be bounded by $\sup_{\pi} \PP^*_{\Epsq} (\phi) \cdot O(H\epsilon_l)$ similarly to the case in Section \ref{appendix:lemma:basic_tot_bound}. For the second term, we proceed as the following:
\begin{align*}
    &\sum_{x_{1:H}\in \Epsq, r_{1:H}} T_{0:H-1} \pi_{1:H} | \mathbf{1}^\top (D^{1,1}_{H:1} - D^{3}_{H:1}) \mathbf{1} | \\
    &\le \underbrace{\sum_{h_H} \sum_{a_H,r_H} \indic{x_{1:H} \in \Epsq} \pi(a_H | h_H) \pi_{1:H-1} T_{0:H-1} \|(D_H^{1,1} - D_H^3) \cdot D_{H-1:1}^{3} \mathbf{1}\|_1}_{(i)} \\
    &\ + \underbrace{\sum_{h_H} \sum_{a_H, r_H} \indic{x_{1:H} \in \Epsq} \pi(a_H | h_H) \pi_{1:H-1} T_{0:H-1} \|D_H^{1,1} \cdot (D_{H-1:1}^{1,1} - D_{H-1:1}^3) \mathbf{1} \|_1}_{(ii)}.
\end{align*}
We first aim to handle the term $(i)$. Here we can divide the cases such that when $x_H \notin \mXt$ and when $x_H \in \mXt$. In the former case, we first note that for any $x \notin \mXt$, we have
$$\|D^1(r|x) - D_+(r|x)\|_1 = \frac{1}{2} |R_1^1(r|x) - R_2^1(r|x)| = \Delta^1(x) \le \Delta(x),$$ 
where the last inequality is followed by the construction of $\mM^1$. From the above, we have
\begin{align*}
    &\sum_{r_H} \indic{x_{1:H} \in \Epsq} \pi(a_H | h_H) \|(D_H^{1,1} - D_H^3) D_{H-1:1}^{3} \mathbf{1} \|_1 \le \Delta(x_H) \indic{x_{1:H} \in \Epsq} \pi(a_H | h_H) \|D_{H-1:1}^{3} \mathbf{1}\|_1.
\end{align*}

In the latter case $x_H \in \mX_{l,3}$, we have $D^{1,1}_H - D^3_H = 0$ by construction. Therefore (i) can be bounded as
\begin{align}
    (i) &\le \sum_{h_H} \pi_{1:H-1} T_{0:H-1} \|D_{H-1:1}^{3} \mathbf{1}\|_1 \sum_{a_H: x_H \notin \mXt} \indic{x_{1:H} \in \Epsq} \Delta(x_H) \pi(a_H | h_H) \nonumber \\
    &\le 2 \sum_{h_H: \{x_{t}\}_{t=1}^{H-1} \cap \mXt \neq \emptyset} \pi_{1:H-1} T_{0:H-1} \PP_{\Epsq}^*(h_H) \epsilon_l \|(D_+)_{H-1:1} \mathbf{1}\|_1 \le 2 \epsilon_l \PP^*_{\Epsq} (\phi).
    \label{eq:tricky_eps3_bound}
\end{align}
The last inequality follows from the fact that since we consider trajectories in $\Epsq$, if $x_H \notin \mXt$, then there is a unique $t^* \le H-1$ such that $D_{t^*}^3 = c_{t^*} \begin{bmatrix} 1 & 0 \\ 0 & -1\end{bmatrix}$ for $c_{t^*}$ that satisfies
\begin{align*}
    c_{t^*} \le \frac{1}{2} |\Delta^1(x_{t^*}) - \Delta^2(x_{t^*})| \le \frac{1}{2} |\Delta(x_{t^*}) - \hat{\Delta}(x_{t^*})| + 2\epsilon_l \le 2\epsilon_l \cdot \frac{\Delta(x_{t^*})}{\Delta(x_H)} \le \frac{2\epsilon_l}{\Delta(x_H)},
\end{align*} 
where the last inequality follows from Corollary \ref{corollary:big_delta_bound}. Furthermore, for all other $t \neq t^* \le H-1$ by construction we have $D_t^3 = (R_+)_t \begin{bmatrix} 1 & 0 \\ 0 & 1\end{bmatrix}$. Therefore we have $\|D^3_{H-1:1} \mathbf{1} \|_1 = c_{t^*} \|(D_+)_{H-1:1} \mathbf{1} \|_1$. We plug this into \eqref{eq:tricky_eps3_bound} and apply the bound on $c_{t^*}$. 

For the second term $(ii)$, we split into two cases when $x_H \notin \mXt$ and $x_H \in \mXt$, or equivalently, $\{x_t\}_{t=1}^{H-1} \cap \mXt \neq \emptyset$ and $\{x_t\}_{t=1}^{H-1} \cap \mXt = \emptyset$. In the former case, we simply reduce $(ii)$ such that
\begin{align*}
    \sum_{r_H} \indic{x_{1:H} \in \Epsq} \|D_H^{1,1} (D^{1,1}_{H-1:1} - D_{H-1:1}^3) \mathbf{1} \|_1 \le \indic{x_{1:H} \in \Epsq} \|(D^{1,1}_{H-1:1} - D_{H-1:1}^3) \mathbf{1} \|_1.
\end{align*}
In order to handle the case $x_H \in \mXt$, let us define $m_t := \max_{t' < t} \Delta(x_{t'})$ for $t \ge 2$ and $m_1 = 1$. Since $D_{H}^{1,1} = (D_-^M)_H = c_{H} \begin{bmatrix} 1 & 0 \\ 0 & -1 \end{bmatrix}$ for some $c_H \le 2 \epsilon_l / m_H$, we have
\begin{align*}
    \sum_{r_H} \indic{x_{1:H} \in \Epsq} \|D_H^{1,1} (D^{1,1}_{H-1:1} - D_{H-1:1}^3) \mathbf{1} \|_1 \le 2\epsilon_l \cdot \frac{\indic{x_{1:H} \in \Epsq}}{m_H} \|(D^{1,1}_{H-1:1} - D_{H-1:1}^3) \mathbf{1}\|_1.
\end{align*}
From the fact above, we can reduce $(ii)$ as the following:
\begin{align*}
    (ii) &\le 2\epsilon_l \sum_{h_H: \{x_t\}_{t=1}^{H-1} \cap \mXt = \emptyset} \pi_{1:H-1} T_{0:H-1} \frac{\PP_{\Epsq}^*(h_H)}{m_H} \|(D^{1,1}_{H-1:1} - D_{H-1:1}^3) \mathbf{1}\|_1 \\
    &\ + \sum_{h_H: \{x_t\}_{t=1}^{H-1} \cap \mXt \neq \emptyset} \pi_{1:H-1} T_{0:H-1} \PP_{\Epsq}^*(h_H) \|(D^{1,1}_{H-1:1} - D_{H-1:1}^3) \mathbf{1}\|_1.
\end{align*}
In order to proceed to the next time step, we need the following recursive relation for any $2 \le t \le H$:
\begin{align}
    &2 \epsilon_l \sum_{h_t: \{x_{t'}\}_{t'=1}^{t-1} \cap \mXt = \emptyset} \pi_{1:t-1} T_{1:t-1} \frac{\PP_{\Epsq}^*(x_{1:t-1}, s_t)}{m_t} \|(D^{1,1}_{t-1:1} - D_{t-1:1}^3) \mathbf{1}\|_1 \nonumber \\
    &\qquad + \sum_{h_t: \{x_{t'}\}_{t'=1}^{t-1} \cap \mXt \neq \emptyset} \pi_{1:t-1} T_{1:t-1} \PP_{\Epsq}^*(x_{1:t-1}, s_t) \|(D^{1,1}_{t-1:1} - D_{t-1:1}^3) \mathbf{1}\|_1 \nonumber  \\
    &\le 4\epsilon_l \PP_{\Epsq}^* (\phi) + 2 \epsilon_l \sum_{h_{t-1}: \{x_{t'}\}_{t=1}^{t-2} \cap \mXt = \emptyset} \pi_{1:t-2} T_{1:t-2} \frac{\PP_{\Epsq}^*(x_{1:t-2}, s_{t-1})}{m_{t-1}} \|(D^{1,1}_{t-2:1} - D_{t-2:1}^3) \mathbf{1}\|_1 \nonumber \\
    &\qquad + \sum_{h_{t-1}: \{x_{t'}\}_{t=1}^{t-2} \cap \mXt \neq \emptyset} \pi_{1:t-2} T_{1:t-2} \PP_{\Epsq}^*(x_{1:t-2}, s_{t-1}) \|(D^{1,1}_{t-2:1} - D_{t-2:1}^3) \mathbf{1}\|_1. \label{eq:eps3_helper_lemma}
\end{align}
By applying \eqref{eq:eps3_helper_lemma} recursively until $t = 2$, we conclude that $(ii) \le 4 H \epsilon_l \cdot \PP_{\Epsq}^* (\phi)$. Plugging this back to \eqref{eq:eps3_three_decomposition} (with a similar bound for the final term), we have 
\begin{align*}
    \sum_{x_{1:H} \in \Epsq, r_{1:H}} \nu(s_1) T_{1:H-1} \pi_{1:H} |\mathbf{1}^\top (D_{H:1}^1 - D_{H:1}^2) \mathbf{1}| \le O(H\epsilon_l) \PP^*_{\Epsq}(\phi). 
\end{align*}
We conclude that $\sum_{\tau:x_{1:H} \in \Epst} |\PP^{1,\pi} (\tau) - \PP^{2,\pi}(\tau) | \le O(H\epsilon_l) \cdot \PP^*_{\Epst} (\phi)$.

\subsubsection{Proof of Equation \eqref{eq:eps3_helper_lemma}}
We start with observing that for $(h_{t-1}, a_{t-1}): \{x_{t'}\}_{t'=1}^{t-1} \cap \mXt = \emptyset$, where we can reduce the sum at time $t$ as
\begin{align*}
    &\sum_{r_{t-1}, s_t} \pi(a_{t-1} | h_{t-1}) T(s_t | s_{t-1},a_{t-1}) \frac{\PP_{\Epsq}^*(x_{1:t-1}, s_t)}{m_t} \| (D_{t-1:1}^{1,1} - D_{t-1:1}^3) \mathbf{1} \|_1 \\
    &\le \sum_{r_{t-1}, s_t} \pi(a_{t-1} | h_{t-1}) T(s_t | s_{t-1},a_{t-1}) \frac{\PP_{\Epsq}^*(x_{1:t-1}, s_t)}{m_t} \|D_{t-1}^{1,1} \cdot (D_{t-2:1}^{1,1} - D_{t-2:1}^3) \mathbf{1} \|_1 \\
    &\quad + \sum_{r_{t-1}, s_t} \pi(a_{t-1} | h_{t-1}) T(s_t | s_{t-1},a_{t-1}) \frac{\PP_{\Epsq}^*(x_{1:t-1}, s_t)}{m_t} \|(D_{t-1}^{1,1} - D_{t-1}^3) \cdot D_{t-2:1}^3 \mathbf{1} \|_1 \\
    &\le \sum_{s_t} \pi(a_{t-1} | h_{t-1}) T(s_t | s_{t-1},a_{t-1}) \frac{\PP_{\Epsq}^*(x_{1:t-1}, s_t)}{m_{t-1}} \|(D_{t-2:1}^{1,1} - D_{t-2:1}^3) \mathbf{1} \|_1 \\
    &\quad + \sum_{s_t} \pi(a_{t-1} | h_{t-1}) T(s_t | s_{t-1},a_{t-1}) \PP_{\Epsq}^*(x_{1:t-1}, s_t) \|D^3_{t-2:1} \mathbf{1} \|_1.
\end{align*}
where in the last inequality we used $m_t \ge m_{t-1}$ by definition. Note that $D^3_{t-2:1} = (D_+)_{t-2:1}$ by the construction of $D^3$. 

For the case $(h_{t-1}, a_{t-1}): \{x_{t'}\}_{t'=1}^{t-1} \cap \mXt \neq \emptyset$, we consider two cases $x_{t-1} \in \mXt$ and $x_{t-1} \notin \mXt$. In the former, we have the remaining terms for $h_{t-1}: \{x_{t'}\}_{t'=1}^{t-2} \cap \mXt = \emptyset$. That is, 
\begin{align*}
    &\sum_{r_{t-1}, s_t} \pi(a_{t-1} | h_{t-1}) T(s_t | s_{t-1},a_{t-1}) \PP_{\Epsq}^*(x_{1:t-1}, s_t) \|(D_{t-1:1}^{1,1} - D_{t-1:1}^3) \mathbf{1}\|_1 \\
    &\le \sum_{r_{t-1}, s_t} \pi(a_{t-1} | h_{t-1}) T(s_t | s_{t-1},a_{t-1}) \PP_{\Epsq}^*(x_{1:t-1}, s_t) \|D_{t-1}^{1,1} \cdot (D_{t-2:1}^{1,1} - D_{t-2:1}^3) \mathbf{1} \|_1 \\
    &\le 2\epsilon_l \sum_{s_t} \pi(a_{t-1} | h_{t-1}) T(s_t | s_{t-1},a_{t-1}) \frac{\PP_{\Epsq}^*(x_{1:t-1}, s_t)}{m_{t-1}} \|(D_{t-2:1}^{1,1} - D_{t-2:1}^3) \mathbf{1} \|_1,
\end{align*}
where the last inequality can be derived as we did in \eqref{eq:tricky_eps3_bound}. 

In the latter case when $x_{t-1} \notin \mXt$, we first note that $\{x_{t'}\}_{t'=1}^{t-2} \cap \mXt \neq \emptyset$. In this case we proceed as the following:
\begin{align*}
    &\sum_{r_{t-1}, s_t} \pi(a_{t-1} | h_{t-1}) T(s_t | s_{t-1},a_{t-1}) \PP_{\Epsq}^*(x_{1:t-1}, s_t) \|(D_{t-1:1}^{1,1} - D_{t-1:1}^3) \mathbf{1}\|_1 \\
    &\le \sum_{r_{t-1}, s_t} \pi(a_{t-1} | h_{t-1}) T(s_t | s_{t-1},a_{t-1}) \PP_{\Epsq}^*(x_{1:t-1}, s_t) \|D_{t-1}^{1,1} \cdot (D_{t-2:1}^{1,1} - D_{t-2:1}^3) \mathbf{1} \|_1 \\
    &\quad + \sum_{r_{t-1}, s_t} \pi(a_{t-1} | h_{t-1}) T(s_t | s_{t-1},a_{t-1}) \PP_{\Epsq}^*(x_{1:t-1}, s_t) \|(D_{t-1}^{1,1} - D_{t-1}^3) \cdot D_{t-2:1}^3 \mathbf{1} \|_1 \\
    &\le \sum_{s_t} \pi(a_{t-1} | h_{t-1}) T(s_t | s_{t-1},a_{t-1}) \PP_{\Epsq}^*(x_{1:t-1}, s_t) \|(D_{t-2:1}^{1,1} - D_{t-2:1}^3) \mathbf{1} \|_1 \\
    &\quad + \sum_{s_t} \pi(a_{t-1} | h_{t-1}) T(s_t | s_{t-1},a_{t-1}) \PP_{\Epsq}^*(x_{1:t-1}, s_t) \Delta(x_{t-1}) \|D_{t-2:1}^3 \mathbf{1} \|_1.
\end{align*}
Using the similar trick as in \eqref{eq:tricky_eps3_bound}, we can reduce $\Delta(x_{t-1}) \|D_{t-2:1}^3 \mathbf{1} \|_1 \le 2\epsilon_l \|(D_+)_{t-2:1} \mathbf{1} \|_1$ since in this trajectory there must exist $t^* \le t-2$ such that $x_{t^*} \in \mXt$. Now we need to combining all terms and sum over $h_{t-1}$:
\begin{align*}
    &2 \epsilon_l \sum_{h_t: \{x_{t'}\}_{t'=1}^{t-1} \cap \mXt = \emptyset} \pi_{1:t-1} T_{1:t-1} \frac{\PP_{\Epsq}^*(x_{1:t-1}, s_t)}{m_t} \|(D^{1,1}_{t-1:1} - D_{t-1:1}^3) \mathbf{1}\|_1 \\
    &\qquad + \sum_{h_t: \{x_{t'}\}_{t'=1}^{t-1} \cap \mXt \neq \emptyset} \pi_{1:t-1} T_{1:t-1} \PP_{\Epsq}^*(x_{1:t-1}, s_t) \|(D^{1,1}_{t-1:1} - D_{t-1:1}^3) \mathbf{1}\|_1 \\
    &\le 2 \epsilon_l \sum_{h_{t-1}: \{x_{t'}\}_{t'=1}^{t-2} \cap \mXt = \emptyset} \pi_{1:t-2} T_{1:t-2} \|(D^{1,1}_{t-2:1} - D_{t-2:1}^3) \mathbf{1}\|_1 \frac{\PP_{\Epsq}^*(x_{1:t-2}, s_{t-1})}{m_{t-1}} \\
    &\ + \sum_{h_{t-1}: \{x_{t'}\}_{t'=1}^{t-2} \cap \mXt \neq \emptyset} \pi_{1:t-2} T_{1:t-2} \|(D^{1,1}_{t-2:1} - D_{t-2:1}^3) \mathbf{1}\|_1 \PP_{\Epsq}^*(x_{1:t-2}, s_{t-1}) \\
    &\ + 2 \epsilon_l \sum_{h_{t-1}} \pi_{1:t-2} T_{1:t-2} \|(D_{+})_{t-2:1} \mathbf{1} \|_1 \PP_{\Epsq}^*(x_{1:t-2}, s_{t-1}).
\end{align*}
The last term is less than $4\epsilon_l \PP_{\Epsq}^*(\phi)$, which proves equation \eqref{eq:eps3_helper_lemma}.

\subsection{Analysis for Case II: $\Epsd$ (equation \eqref{eq:eps_2_bound})}
\label{appendix:complete_eps2}
For any $(x_i,x_j) \in \mX_l$, let $\PP^c(r_j | x_j, r_i, x_i) := \mu^1(x_i, x_j) / R_+(r_i | x_i)$ for $c = 1,2$ (recall that we use $R_+(r|x) := \frac{1}{2}(R_1^1 + R_2^1) (r|x)$, and $R_-^c (r|x) := \frac{1}{2} (R_1^c - R_2^c) (r|x)$). We first observe that
\begin{align}
    \| \PP^1(r_j | x_j, r_i, x_i) - \PP^2(r_j | x_j, r_i, x_i) \|_1 \le O(\epsilon_l) / \aR (r_i | x_i), \label{eq:eps2_conditional_tv_distance}
\end{align}
since from the Bayes' rule, we have
\begin{align*}
    \PP^c(r_j | x_j, r_i, x_i) &= \frac{R_1^c(r_i|x_i)}{R_1^c(r_i|x_i) + R_2^c(r_i|x_i)} R_1^c(r_j|x_j) + \frac{R_2^c(r_i|x_i)}{R_1^c(r_i|x_i) + R_2^c(r_i|x_i)} R_2^c(r_j|x_j), \\
    &= \frac{\PP^c (r_i, r_j | x_i, x_j)}{R_+(r_i|x_i)}, 
\end{align*} 
for $c = 1,2$. We start by writing the target errors as usual.
\begin{align*}
    \sum_{\tau: x_{1:H} \in \Epsd} &| \PP^{1,\pi}(\tau) - \PP^{2,\pi}(\tau) | = \sum_{(x,r)_{1:H}} \indic{x_{1:H} \in \Epsd} T_{0:H-1} \pi_{1:H} \cdot |(\PP^1 - \PP^2) (r_{1:H} | x_{1:H}) | \\
    &= \sum_{h_H} \pi_{1:H-1} T_{0:H-1} \sum_{a_H, r_H} \indic{x_{1:H} \in \Epsd} \pi(a_H | h_H) \cdot | (\PP^1 - \PP^2) (r_{1:H} | x_{1:H}) |,
\end{align*}
where we define 
\begin{align*}
    \PP^c (r_{1:t} | x_{1:t}) := \frac{1}{2} \sum_{m=1}^2 \Pi_{t'=1}^t R^c_m(r_{t'} | x_{t'}).
\end{align*}
for $c = 1,2$ and any $t \in [H]$. Note that by Lemma \ref{lemma:eps2_count_distinguishable}, we are guaranteed that all trajectories in $\Epsd$ have at most 2 $\delta_l$-distinguishable state-actions in both models $\mM^1, \mM^2$. We split the cases when the number of distinguishable state-actions until time $H-1$ is whether $d = 2, 1, 0$. Let $\mX_{d,t}$ be a set of length $t$ transition histories such that the number of distinguishable state-actions is exactly $d$. That is,
\begin{align}
    \label{eq:helper_def_history_eps2}
    \mX_{d,t} = \left\{x_{1:t-1} \Big| \sum_{t' = 1}^{t-1} \indic{\Delta^1(x_{t'}) > 0 \text{ or } \Delta^2(x_{t'}) > 0} = d \right\}.
\end{align}
With a slight abuse in notation, we overwrite the notation $\Delta(x) := \max(\Delta^1(x), \Delta^2(x))$. Note that when $\Delta(x) = 0$ we have $D^1(\cdot|x) = D^2(\cdot|x)$ by construction of models $\mM^1$ and $\mM^2$.

We start from length $H$ histories $h_H$. We divide cases into whether $h_H$ belongs to $\mX_{2,H}$, $\mX_{1,H}$ or $\mX_{0,H}$.

\paragraph{Case i: $h_H: x_{1:H-1} \in \mX_{2,H}$.} Since we consider trajectories only in $\Epsd$, we have $\Delta (x_H) = 0$ and therefore the probability of observing $r_H$ under any history $h_H$ is the same, {\it i.e.,} $(\PP^1 - \PP^2) (r_H | h_H) = 0$. Hence we have
\begin{align*}
    &\sum_{a_H, r_H} \indic{x_{1:H} \in \Epsd} \pi(a_H | h_H) \cdot |(\PP^1 - \PP^2) (r_{H:1} | x_{H:1})| \\
    &= \sum_{a_H} \indic{x_{1:H} \in \Epsd} \pi(a_H | h_H) \cdot | (\PP^1 - \PP^2)(r_{H-1:1} | x_{H-1:1}) |.
\end{align*}

\paragraph{Case ii: $h_H: x_{1:H-1} \in \mX_{1,H}$.} We consider two cases: when $\Delta(x_H) = 0$ or $\Delta(x_H) > 0$. Let us define an event $\Eps_{d,t}$ which is defined as the following:
\begin{align}
    \label{eq:helper_def_trajectory_eps2}
    \Eps_{d, t} = \Epsd \cap \{\tau = (x,r)_{1:H} | x_{1:t-1} \in \mX_{d-1,t} \cap \Delta(x_t) > 0 \},
\end{align} 
for $d = 2,1$ and $t \in [H]$. Now in case $\Delta(x_H) = 0$, we again have $(\PP^1 - \PP^2) (r_H | h_H) = 0$. Therefore we have
\begin{align*}
    \sum_{r_H} | (\PP^1-\PP^2)(r_{H:1}|x_{H:1}) | = | (\PP^1-\PP^2)(r_{H-1:1}|x_{H-1:1}) |.
\end{align*}
In order to handle the latter case when $\Delta(x_H) > 0$, we first define $t^*$ and $p_t$ as the following:
\begin{align}
    \label{def:p_t}
    t^*&: t^* < t, \ s.t. \ \Delta(x_{t^*}) > 0, \nonumber \\
    p_t &:= \PP^1 (r_{t^*} | x_{t^*}) = \PP^2 (r_{t^*} | x_{t^*}) = \aR (r_{t^*} | x_{t_t^*}).
\end{align} 
Note that with the above definition, for length $t$ history $h_t$ in any trajectories in $\Epsd$, we have
\begin{align*}
    \PP^c(r_{t:1} | x_{t:1}) = \PP^c(r_t | x_t, x_{t^*}, r_{t^*}) \cdot \PP^c( r_{t-1:1} | x_{t-1:1}), \qquad c = 1,2,
\end{align*}
since for any $t' \neq t^*, t$, we have $\Delta^1(x_{t'}) = \Delta^2(x_{t'}) = 0$. Recall the inequality \eqref{eq:eps2_conditional_tv_distance},
\begin{align*}
    &\sum_{r_H} \indic{x_{1:H} \in \Epsd} \pi(a_H | h_H) | (\PP^1-\PP^2)(r_{H:1}|x_{H:1}) | \\
    &\le \sum_{r_H} \indic{x_{1:H} \in \Epsd} \pi(a_H | h_H) |(\PP^1 - \PP^2) (r_H | h_H) | \PP^1(r_{1:H-1} | x_{1:H-1}) \\
    &+ \sum_{r_H} \indic{x_{1:H} \in \Epsd} \pi(a_H | h_H) \PP^2 (r_H | h_H) |(\PP^1 - \PP^2) (r_{1:H-1}|x_{1:H-1}) | \\
    &\le \indic{x_{1:H} \in \Eps_{2,H}} \pi(a_H | h_H) \left(\frac{\epsilon_l}{p_H} \PP^1 (r_{1:H-1}|x_{1:H-1}) + | (\PP^1 - \PP^2) (r_{1:H-1} | x_{1:H-1}) | \right).
\end{align*}
In the last inequality, since we are handling the special case when $x_{1:H} \in \Eps_{2,H}$, we replace $\indic{x_{1:H} \in \Epsd}$ by $\indic{x_{1:H} \in \Eps_{2,H}}$. 

\paragraph{Case iii: $h_H \in \Eps_{0,H}$.} We first observe that $\PP^1(r_{1:H-1} | x_{1:H-1}) = \PP^2(r_{1:H-1} | x_{1:H-1})$. Furthermore, in this case $\PP^1(r_H | h_H) = \PP^1 (r_H | x_{H}) = R_+(r_H | x_H)$. Thus
\begin{align*}
    \sum_{a_H, r_H} \indic{x_{1:H} \in \Epsd} \pi(a_H | h_H) | (\PP^1-\PP^2) (r_{H:1} | x_{H:1}) | = 0.
\end{align*}

\paragraph{Combine all cases.} Plugging all of this into the summation over $h_H$:
\begin{align}
    \label{eq:eps2_induction_bound}
    &\sum_{h_H} \pi_{1:H-1} T_{0:H-1} \sum_{a_H, r_H} \indic{x_{1:H} \in \Epsd} \pi(a_H | h_H) | (\PP^1-\PP^2) (r_{H:1} | x_{H:1}) | \nonumber \\
    &\le \sum_{h_H} \pi_{1:H-1} T_{0:H-1} | (\PP^1 - \PP^2) (r_{H-1:1} | x_{H-1:1}) | \PP_{\Epsd}^* (h_H) \nonumber \\
    &+ \sum_{h_H: x_{1:H-1} \in \mX_{1,H}} \pi_{1:H-1} T_{0:H-1} \PP^1(r_{H-1:1}|x_{H-1:1})  \frac{\epsilon_l}{p_H} \sum_{a_H} \indic{x_{1:H} \in \Eps_{2,H}} \pi(a_H | h_H).
\end{align}
Finally, we control the last term in the above equation. We again consider the case when $\Delta(x_{H-1}) = 0$ and $\Delta(x_{H-1}) > 0$ which gives
\begin{align}
    \label{eq:eps2_tricky_bound}
    &\sum_{h_H: x_{1:H-1} \in \mX_{1,H}} \pi_{1:H-1} T_{0:H-1} \PP^1(r_{H-1:1} | x_{H-1:1})  \frac{\epsilon_l}{p_H} \sum_{a_H} \PP^*_{\Eps_{2,H}} (x_{1:H}) \pi(a_H | h_H) \nonumber \\
    &\le \sum_{h_{H-1}: x_{1:H-2} \in \mX_{1,H-1}} \pi_{1:H-2} T_{0:H-2} \PP^1(r_{H-2:1} | x_{H-2:1})  \frac{\epsilon_l \PP^*_{\Eps_{2,H}} (h_{H-1})}{p_{H-1}} \nonumber \\
    &+ 2\epsilon_l \sum_{h_{H-1}: x_{1:H-2} \in \mX_{0,H-1}} \pi_{1:H-2} T_{0:H-2} \PP^1(r_{H-2:1} | x_{H-2:1})  \sum_{a_{H-1}: \Delta(x_{H-1}) > 0} \pi(a_{H-1}|h_{H-1}) \PP^*_{\Eps_{2,H}} (x_{1:H-1}) \nonumber \\
    &\le \sum_{h_{H-1}: x_{1:H-2} \in \mX_{1,H-1}} \pi_{1:H-2} T_{0:H-2} \PP^1(r_{H-2:1} | x_{H-2:1})  \frac{\epsilon_l \PP^*_{\Eps_{2,H}} (h_{H-1})}{p_{H-1}} \nonumber \\
    &+ 2\epsilon_l \sum_{h_{H-1}: x_{1:H-2} \in \mX_{0,H-1}}   \sum_{a_{H-1}: \Delta(x_{H-1}) > 0} \PP^{1, \pi} (h_{H-1}, a_{H-1}) \PP_{\Eps_{2,H}}^*(x_{1:H-1}).
\end{align}
The first term comes from the fact that $p_{H-1} = p_H$ in case $\Delta(x_{H-1}) = 0$. For the second term, we note that since $x_{1:H-2} \in \mX_{0,H-1}$, we have $\PP^1(r_{H-2:1} | x_{H-2:1}) = (R_+)_{H-2:1}$, which is equivalently $\PP^1(r_{H-1:1} | x_{H-1:1}) = (R_+)_{H-1} \PP^1(r_{H-2:1} | x_{H-2:1})$. We apply this relation to obtain the above inequality \eqref{eq:eps2_tricky_bound}. 

We repeat this induction recursively until time step 1, and we get
\begin{align*}
    &\sum_{h_H: x_{1:H-1} \in \mX_{1,H}} \pi_{1:H-1} T_{0:H-1} \PP^1(r_{H-1:1} | x_{H-1:1}) \frac{\epsilon_l \PP^*_{\Eps_{2,H}} (h_H)}{p_H} \\
    &\le 2 \epsilon_l \sum_{t=1}^{H-1} \sum_{h_t: x_{1:t-1} \in \mX_{0,t}} \sum_{a_t: \Delta(x_t) > 0} \PP_{\Eps_{2,H}}^* (x_{1:t}) \PP^{\pi} (h_t,a_t) \\
    &\le 2 \epsilon_l \sum_{t=1}^{H-1} \sum_{h_t: x_{1:t-1} \in \mX_{0,t}} \sum_{a_t: \Delta(x_t) > 0} \PP_{\Eps_{1,t}}^* (x_{1:t}) \PP^{\pi} (h_t,a_t).
\end{align*}
The last inequality follows from the fact that conditioned on $x_{1:t}$ such that $x_{1:t-1} \in \mX_{0,t}$ and $\Delta(x_t) > 0$, we get $\PP_{\Eps_{1,t}}^*(x_{1:t}) \ge \PP_{\Eps_{2,H}}^*(x_{1:t})$. 

Finally, we observe that $\Eps_{1,t} \cap \Eps_{1,t'} = \emptyset$ for all $t,t' \in [H]$ such that $t \neq t'$, and $\cup_{t=1}^H \Eps_{1,t} \subseteq \Epsd$. Therefore there is no duplication of histories in the summation and we can safely proceed to:
\begin{align*}
    &\sup_{\pi} \sum_{t=1}^{H-1} \sum_{h_t: x_{1:t-1} \in \mX_{0,t}} \sum_{a_t: \Delta(x_t) > 0} \PP_{\Eps_{1,t}}^* (h_{t},a_t) \PP^{\pi} (h_t,a_t) \\
    &= \sup_{\pi} \sum_{t=1}^{H-1} \sum_{(h_t,a_t): (x_{1:t-1} \in \mX_{0,t}) \cap \Delta(x_t) > 0} \PP^{\pi} (\Eps_{1,t} | h_t,a_t) \PP^{\pi} (h_t,a_t) \\
    &\le \sup_{\pi} \sum_{t=1}^{H-1} \sum_{(h_t,a_t): (x_{1:t-1} \in \mX_{0,t}) \cap \Delta(x_t) > 0} \PP^\pi (\Eps_{1,t}, h_t, a_t) \le \sup_{\pi} \sum_{t=1}^{H-1} \PP^\pi (\Eps_{1,t}) \le \sup_{\pi} \PP^{\pi} (\Epsd).  
\end{align*}
This concludes that 
\begin{align*}
    \sum_{h_H \in \Eps_{1,H}} &\pi_{1:H-1} T_{0:H-1} \PP^1(r_{H-1:1} | x_{H-1:1})  \frac{\epsilon_l \PP^*_{\Epsd} (h_H)}{p_H} \le 2\epsilon_l \PP_{\Epsd}^{*} (\phi),
\end{align*}
and we can plug this back to \eqref{eq:eps2_induction_bound} and apply the same argument recursively until time step drops down to 1. This gives
\begin{align*}
    \sum_{\tau: x_{1:H} \in \Epsd} | \PP^{1,\pi} (\tau) - \PP^{2,\pi} (\tau) | \le O(H \epsilon_l) \PP_{\Epsd}^* (\phi).
\end{align*}

\newpage

\section{Deferred Analysis in Section \ref{section:main_algorithm}}
\label{Appendix:complete_upper_bound}

\subsection{Formulation of the LP for Model Recovery}
\label{appendix:formulation_lp}
We describe a detailed LP formulation to obtain an empirical reward models. Let $c_u(x_i, x_j)$ and $c_l(x_i, x_j)$ are upper and lower confidence bounds for $u(x_i, x_j)$ respectively:
\begin{align*}
    c_u(x_i, x_j) &:= \hat{\mu} (x_i, x_j) - \hat{p}_+ (x_i) \hat{p}_+ (x_j) + \sqrt{\frac{\iota_2}{n(x_i, x_j)}} + \epsilon_0^2, \\
    c_l(x_i, x_j) &:= \hat{\mu} (x_i, x_j) - \hat{p}_+ (x_i) \hat{p}_+ (x_j) - \sqrt{\frac{\iota_2}{n(x_i, x_j)}} - \epsilon_0^2.
\end{align*}
We can write down the linear program to find variables $\hat{l}(x)$ with several linear constraints. The first constraint is upper bound on the multiplication of two reward differences:
\begin{align}
    \label{eq:constrinats_upper_bound}
    \hat{l}(x_i) + \hat{l}(x_j) \le \log \left( |c_u(x_i,x_j)| \vee |c_l(x_i,x_j)| \right), &\ \forall x_i, x_j \in \mS \times \mA, 
\end{align}
The second constraint is, for all $x_i, x_j \in \mS\times\mA$ such that signs of both upper and lower bounds are equal, a lower confidence bound:
\begin{align}
    \label{eq:constrinats_lower_bound}
    \hat{l}(x_i) + \hat{l}(x_j) &\ge \log \left( |c_u(x_i,x_j)| \wedge |c_l(x_i,x_j)| \right), \ \forall (x_i, x_j): \ c_u(x_i, x_j) \cdot c_l(x_i, x_j) > 0.
\end{align}
Finally, we add a regularity condition for differences in probabilities:
\begin{align}
    \label{eq:constraints_regularity}
    10 \log(\epsilon_0) \le \hat{l}(x) \le \log \left( \left(\hat{p}_+ (x) \wedge 1 - \hat{p}_+(x) \right) + \sqrt{\iota_1 / n(x)} + \epsilon_0^2 \right), &\ \forall x \in \mS \times \mA.
\end{align}
with a properly set confidence parameter $\iota_1 = O(\log (SA/\eta))$. Now we can solve the LP feasibility problem for $\hat{l}(x)$ and find parameters for an empirical model. The LP formulation can be found in \eqref{eq:LP_formulation}. The minimization objective $\sum \hat{l}(x)$ encourages to ignore small differences in two reward models at each state-action. 
\begin{algorithm}[t]
    \caption{Reward Model Recovery from Second-Order Correlations}
    \label{algo:LMDP_recovery_detail}
    \begin{algorithmic}[1]
        \STATE Solve an LP for $\hat{l}(x)$ for all $x \in \mS \times \mA$ such \begin{align}
            \min_{\hat{l}(x): x \in \mS \times \mA} & \qquad \sum_{x \in \mS \times \mA} \hat{l}(x), \nonumber \\
            \text{s.t.} & \qquad \hat{l}(x) \text{ satisfy } \eqref{eq:constrinats_upper_bound}, \eqref{eq:constrinats_lower_bound}, \text{ and } \eqref{eq:constraints_regularity} \label{eq:LP_formulation}
        \end{align} 
        \STATE Decide signs $sign(x)$ that satisfies \eqref{eq:constraints_sign}. 
        \STATE Clip $\hat{l}(x)$ within $(-\infty, u(x)]$ where $u(x) = \log \left(\min(\hat{p}_+(x), 1 - \hat{p}_+(x) \right)$ for all $x$.
        \STATE Set $\hat{R}_-(r=1|x) = sign(x) \cdot \exp(\hat{l}(x))$ and $\hat{R}_-(r=0|x) = -sign(x) \cdot \exp(\hat{l}(x))$ for all $x$.
        \STATE Let $\hat{R}_1(r|x) = \hat{R}_+(r|x) + \hat{R}_-(r|x), \hat{R}_2(r|x) = \hat{R}_+(r|x) - \hat{R}_-(r|x)$ for all $r,x$.
        \STATE Return $\hat{\mM} = (\mS, \mA, \hat{T}, \hat{\nu}, \{\hat{R}_m\}_{m=1}^2)$.
    \end{algorithmic}
\end{algorithm}

After we obtain a value for $l(x) = \log |\hat{\Delta}(x)|$, it remains to find signs of $\hat{p}_-(x)$. We find a correct assignment for $sign(x)$ with the following constraints:
\begin{align}
    \label{eq:constraints_sign}
    sign(x_i) sign(x_j) &= sign(\hat{u}(x_i,x_j)), \ \forall (x_i, x_j): \ c_u(x_i, x_j) \cdot c_l(x_i, x_j) > 0.
\end{align}
Solving constraints \eqref{eq:constraints_sign} can be formulated as 2-Satisfiability (2-SAT) problem \cite{aspvall1979linear} and thus can be efficiently solved. The full algorithm is described in Algorithm \ref{algo:LMDP_recovery_detail}.

\subsection{Proof of Lemma \ref{lemma:feasible_optimism}}
\label{appendix:optimism_feasibility}
\paragraph{Magnitude Constraints} We start with confidence bounds for estimated probabilities that holds with probabilities at least $1 - \eta$:
\begin{align*}
    | p_+(x) - \hat{p}_+ (x) | &\le \sqrt{\frac{c_1 \log(SA/\eta)}{n(x)}}, & \qquad \forall x \in \mS \times \mA, \\
    | \mu(x_i, x_j) - \hat{\mu}(x_i, x_j) | &\le \sqrt{\frac{c_2 \log(SA/\eta)}{n(x_i, x_j)}}, & \qquad \forall x_i, x_j \in \mS \times \mA,
\end{align*}
for some absolute constants $c_1, c_2$. Furthermore, by construction the number of visit counts for any pair is less than the visit counts of a single state-action: 
\begin{align*}
    \min( n(x_i), n(x_j) ) \ge n(x_i, x_j), & \qquad \forall x_i, x_j \in \mS \times \mA.
\end{align*}
Then we consider a model $\mM^1$ that approximates the true model with non-zero probability for every reward:
\begin{align*}
    &R_m^1(r|x) = R_m(r|x), \ m \in \{1,2\}, &\qquad r \in \{0,1\}, x \in \mS \times \mA: |\Delta(x)| > \epsilon_0^5, \\
    &\begin{cases} R_1^1(r|x) = \epsilon_0^3 + (1 - 2\epsilon_0^3) R_1(r|x) + (2r-1) \epsilon_0^4 \\ R_2^1(r|x) = \epsilon_0^3 + (1 - 2\epsilon_0^3) R_2(r|x) - (2r-1) \epsilon_0^4 \end{cases} , &\qquad r \in \{0,1\}, x \in \mS \times \mA: |\Delta(x)| \le \epsilon_0^5,
\end{align*}
For this model, we have
\begin{align*}
    |(R_+^1 - R_+) (r|x) | &\le \epsilon_0^3, &\qquad \forall r \in \{0,1\}, x \in \mS \times \mA, \\
    |(\PP^1 - \PP) (r_i, r_j | x_i, x_j) | &\le 2 \epsilon_0^3, &\qquad  \forall r_i, r_j \in \{0,1\}, x_i, x_j \in \mS \times \mA, \\
    \Delta^1(x) &\ge \epsilon_0^5, &\qquad \forall x \in \mS \times \mA.
\end{align*}
We show that one feasible solution $l^*(x)$ can be obtained from $\mM^1$ 
\begin{align*}
    l^*(x) := \log \left| p_-^1 (x) \right| = \log (\Delta^1(x)).
\end{align*}
We first check the first constraint \eqref{eq:constrinats_upper_bound}:
\begin{align*}
    \Delta^1(x_i) \Delta^1(x_j) &= |\mu^1(x_i, x_j) - p_+^1(x_i) p_+^1(x_j)| \\
    &\le |\hat{\mu}(x_i, x_j) - \hat{p}_+(x_i) \hat{p}_+(x_j)| + \sqrt{\frac{c_2 \log(SA/\eta)}{n(x_i, x_j)}} + \epsilon_0^2 \\
    &= \max( |c_u(x_i, x_j)|, |c_l(x_i, x_j)| ).
\end{align*}
For the second constraint \eqref{eq:constrinats_lower_bound} follows similarly when $c_u(x_i, x_j) \cdot c_l(x_i, x_j) > 0$. In the other case, $\Delta^1(x_i) \Delta^1(x_j)$ can be as small as 0, which implies that $l(x_i) + l(x_j)$ is unbounded below. For the final constraint \eqref{eq:constraints_regularity}, we observe that
\begin{align*}
    \epsilon_0^{5} \le \Delta^1(x) &\le p_+^1(x) \le \hat{p}_+ (x) + \sqrt{\frac{c_1 \log(SA/\eta)}{n(x)}} + \epsilon_0^3, &\qquad \forall x \in \mS \times \mA. 
\end{align*}
Thus $l^*(x)$ satisfies all magnitude constraints \eqref{eq:constrinats_upper_bound}, \eqref{eq:constrinats_lower_bound}, and \eqref{eq:constraints_regularity}.

\paragraph{Sign Constraint} For the sign constraint \eqref{eq:constraints_sign}, we note that whenever $c_u(x_i, x_j) \cdot c_l(x_i, x_j) > 0$, we have
\begin{align}
    sign(p_-^1(x_i) \cdot p_-^1(x_j)) = sign(x_i) sign(x_j) = sign(c_u(x_i, x_j)), \label{eq:sign_equality}
\end{align}
which can be inferred by
\begin{align*}
    p_-^1(x_i) p_-^1(x_2) &= \PP^1(r_i = 1, r_j = 1 | x_i, x_j) - p_+^1(x_i) p_+^1(x_j) \\
    &= \hat{\PP}(r_i = 1, r_j = 1 | x_i, x_j) - \hat{p}_+(x_i) \hat{p}_+(x_j) + errors,
\end{align*}
where $errors$ is in order of $O \left(\sqrt{c_2 \log(SA/\eta) / n(x_i, x_j)} + \epsilon^2 \right)$. This implies that 
\begin{align*}
    c_l(x_i, x_j) \le p_-^1(x_i) p_-^1(x_j) \le c_u(x_i, x_j),
\end{align*}
which concludes that reward models from $\mM^1$ satisfy \eqref{eq:constraints_sign}.

\subsection{Proof of Lemma \ref{lemma:reward_free_guarantee}}
\label{appendix:reward_free_guarantee}
At this point, we no longer assume that the true transition and initial state models are given, but instead we use estimated quantities $\hat{T}, \hat{\nu}$ from Algorithm \ref{algo:pure_explore}. 

\paragraph{Notations} Let us define a few variables to proceed. We denote $\#_k(x)$ as visit counts at $x$ during the $k^{th}$ episode. A similar quantity $\#_k(x_i, x_j)$ is 1 if data for a pair $(x_i, x_j)$ is collected in the $k^{th}$ episode and 0 otherwise. Let $\wpi_k$ be the policy executed in the $k^{th}$ episode. Let $n_k(x) := \sum_{k'=1}^k \#_{k'}(x)$, $n_k(x_i, x_j) = \sum_{k'=1}^k \#_{k'}(x_i, x_j)$ and the expected quantities $\bar{n}_k (x) = \sum_{k'=1}^k \Exs^{\wpi_{k'}} [\#_{k'} (x)]$, $\bar{n}_{k'=1}^k (x_i, x_j) = \sum_{k'=1}^k \Exs^{\wpi_{k'}} [\#_{k'} (x_i, x_j)]$. We define a desired high probability event $\Eps_{pe}$ for martingale sums:
\begin{align}
    \label{eq:def_martingale_event}
    n_k(x) \ge \frac{1}{2} \bar{n}_k (x) - c_l \log (K/\eta), &\qquad \forall k \in [K], x \in \mS \times \mA, \nonumber \\
    n_k(x_i, x_j) \ge \frac{1}{2} \bar{n}_k (x_i, x_j) - c_l \log(K/\eta), &\qquad \forall k \in [K], x_i, x_j \in \mS \times \mA,
\end{align}
for some absolute constant $c_l > 0$. With a standard measure of concentration argument for  martingale sums~\cite{wainwright2019high}, we can show that $\PP(\Eps_{pe}) \ge 1 - \eta$. We also denote $\hat{T}_k, \hat{\nu}_k$ for the empirically estimated transition and initial distribution models at the beginning of $k^{th}$ episode. Let $v_1$ be a $2$-tuple $(null, null)$ and $i_1 = 1$.

\paragraph{Number of Episodes $K$} We first show that we terminate Algorithm \ref{algo:pure_explore} after at most $K$ episodes with probability at least $1 - \eta$ where
\begin{align*}
    K = C \cdot \frac{S^2 A}{\epsilon_{pe}^2} (H + A) \log (K/\eta),
\end{align*}
for some absolute constant $C > 0$. Let us examine $\widetilde{V}_0$ at the $k^{th}$ episode. This can be decomposed as
\begin{align*}
    \widetilde{V}_0 &= \sqrt{\iota_\nu / k} + \sum_{s} \hat{\nu}_k (s) \cdot \widetilde{V}_1 (i_1, v_1, s) \\
    &\le \sqrt{\iota_\nu / k} + \| \hat{\nu}_k (s) - \nu(s) \|_1 + \sum_{s} \nu(s)  \cdot \widetilde{Q}_1 ((i_1,v_1, s), (a,z)) \\
    &\le 2 \sqrt{\iota_\nu / k} + \Exs^{\wpi_{k}} \left[\widetilde{Q}_1 ((i_1,v_1, s_1), \wpi_k(v_1,s_1)) \right].
\end{align*}
Note that $(a_t, z_t) = \wpi_k(v_t, s_t)$ for $t \in [H]$. We can recursively bound expectation of $\widetilde{Q}_t$ for $t \ge 1$:
\begin{align*}
    \Exs^{\wpi_{k}} &\left[\widetilde{Q}_t ((i_t, v_t, s_t), (a_t,z_t)) \right] \\
    &= \Exs^{\wpi_{k}} \left[b_r(i_t, v_t, z_t)  + b_T(s_t,a_t) \right] + \Exs^{\wpi_k} \left[\sum_{s_{t+1}} \hat{T}_k(s_{t+1} | s_t,a_t) \cdot \widetilde{V}_{t+1} (i_{t+1}, v_{t+1}, s_{t+1}) \right] \\
    &\le \Exs^{\wpi_{k}} \left[\sqrt{\frac{\iota_2 \indic{collect}}{n_k (v_{t+1})}} + 2 \sqrt{\frac{\iota_T}{n_k (s_t,a_t)}} \right]  + \Exs^{\wpi_{k}} \left[\widetilde{Q}_{t+1} ((i_{t+1}, v_{t+1}, s_{t+1}), (a_{t+1},z_{t+1})) \right],
\end{align*}
where $\indic{collect}$ is a short hand for $\indic{i_t = 2 \cap z_t = 1}$. We summarize all appended terms to get
\begin{align*}
    \widetilde{V}_0 &\le 2\sqrt{\iota_\nu / k} + \sum_{t=1}^H \Exs^{\wpi_k} \left[ 2 \sqrt{\frac{\iota_T}{n_k (s_t,a_t)}} + \sqrt{\frac{\iota_2 \indic{collect}}{n_k (v_{t+1})}} \right] \\
    &\le 2\sqrt{\iota_\nu / k} + \sum_{(s,a)} 2 \sqrt{\frac{\iota_T}{n_k (s,a)}} \cdot \Exs^{\wpi_k} \left[ \#_k (s,a) \right] + \sum_{(x_i, x_j)} \sqrt{\frac{\iota_2}{n_k (x_i, x_j)}} \cdot \Exs^{\wpi_k} \left[\#_k (x_i, x_j) \right] \\
    &\le 2\sqrt{\iota_\nu / k} + \sum_{x} 2H \cdot \indic{\bar{n}_k (x) < 4 \cdot c_l \log(K/\eta) } \\
    &\quad + 4 \sum_{x} \sqrt{\frac{\iota_T}{\bar{n}_k (x)}} \cdot (\bar{n}_{k+1}(x) - \bar{n}_k (x)) + 4\sum_{(x_i, x_j)} \sqrt{\frac{\iota_2}{\bar{n}_k (x_i, x_j)}} \cdot (\bar{n}_{k+1}(x_i, x_j) - \bar{n}_k (x_i, x_j)).
\end{align*}
We now sum over $K$ episodes, then
\begin{align*}
    K \epsilon_{pe} \le 4 \sqrt{\iota_\nu K} + O(c_l HSA \log(KSA/\eta)) + 8 \sum_{x} \sqrt{\iota_T \bar{n}_{K+1} (x)} + 8 \sum_{(x_i, x_j)} \sqrt{\iota_2 \bar{n}_{K+1} (x_i, x_j)}.
\end{align*}
We now note that $\sum_{x} \bar{n}_{K+1} (x) = HK$ and $\sum_{(x_i, x_j)} \bar{n}_{K+1} (x_i, x_j) \le K$. Using a Cauchy-Schwartz inequality, we get
\begin{align*}
    K \epsilon_{pe} \le O \left(\sqrt{\iota_\nu K} + HSA \log(SA/\eta) + \sqrt{\iota_T HSAK} + \sqrt{\iota_2 S^2 A^2 K} \right). 
\end{align*}
Note we plug our design of confidence interval parameters $\iota_\nu = O(S \log(K/\eta))$, $\iota_T = O(S \log(KSA/\eta))$ and $\iota_2 = O(\log (KSA/\eta))$. This gives a bound that $K \le O\left(S^2 A (A+H) \epsilon_{pe}^{-2} \log(KSA/\eta) \right)$. Hence we terminate Algorithm \ref{algo:pure_explore} after at most $K$ episodes under the event $\Eps_{pe}$
\begin{align*}
    K = C \cdot S^2 A(A+H) \epsilon_{pe}^{-2} \log(HSA/(\epsilon_{pe} \eta)).
\end{align*}
Note that from a concentration of martingale sums, $\Eps_{pe}$ happens with probability at least $1 - \eta$.

\paragraph{Bound on $\PP_{\Eps_l'}^*(\phi)$} We now show the second part of the lemma: under the event $\Eps_{pe}$, we have
\begin{align*}
    \PP_{\Eps_l'}^* (\phi) \le O(H \epsilon_{pe}) \cdot \max \left(1, \sqrt{n_l / \iota_2} \right). 
\end{align*}
We first observe that 
\begin{align*}
    \PP_{\Eps_l'}^* &= \sup_{\pi} \PP^{\pi} (\Eps_l') \le \sup_{\pi} \PP^{\pi} \left(\cup_{t_1=1}^{H-1} \{ \exists t_2 > t_1, \ s.t. \ (x_{t_1}, x_{t_2}) \in \mX_l \cap \mX_{l-1}^c \} \right) \\
    &\le \sup_{\pi} \sum_{t_1=1}^{H-1} \PP^{\pi} \left(\exists t_2 > t_1, \ s.t. \ (x_{t_1}, x_{t_2}) \in \mX_l \cap \mX_{l-1}^c \right).
\end{align*}
Now in the augmented MDP $\tmM$ consider a class of augmented policies $\widetilde{\Pi}_{t_1, l}$ such that for a fixed $t_1$, a policy $\wpi$ picks the first state of a pair only at time step $t_1$, {\it i.e.,} $z_{t} = 0$ for $t < t_1$ and $z_{t_1} = 1$, and then pick the second state of the pair whenever it encounters that makes the pair belong to $\mX_l$, {\it i.e.,} we set $z_{t_2} = 1$ whenever $(x_{t_1}, x_{t_2}) \in \mX_l$ for $t_2 > t_1$. Within this policy class, define $\wQ^1$ and $\widetilde{V}^1$ as the following: 
\begin{align*}
    b_T^1(s,a) &= \left(1 \wedge \sqrt{\frac{\iota_T}{n(s,a)}} \right), \quad b_r^1(i,v,z) = \indic{i = 2 \cap z=1 \cap (v' \in \mX_l \cap \mX_{l-1}^c) } \cdot \left(1 \wedge \sqrt{\frac{\iota_2}{n_l}} \right),
\end{align*}
\begin{align*}
    \wQ_t^1 ((i,v,s), (a,z)) = 1 \wedge \left( b_r + \sum_{s'} \hat{T}(s' | s,a) \cdot \widetilde{V}_{t+1}^1 (i', v',s') + b_T \right), 
\end{align*}
with $\widetilde{Q}_{H+1}^1 = 0$. For $\widetilde{V}^1$, we define it as,
\begin{align*}
    \widetilde{V}_{t}^1 (i,v,s) &= \max_{(a,z) \in \widetilde{\mA}} \widetilde{Q}_t^1((i,v,s),(a,z)), & \text{if } t > t_1, \\
    \widetilde{V}_{t}^1 (i,v,s) &= \max_{a \in \mA} \widetilde{Q}_t^1((i,v,s),(a,1)), & \text{if } t = t_1, \\
    \widetilde{V}_{t}^1 (i,v,s) &= \max_{a \in \mA} \widetilde{Q}_t^1((i,v,s),(a,0)), & \text{if } t < t_1,
\end{align*}
and $\widetilde{V}_0^1 = \sqrt{\iota_\nu / K} + \sum_s \nu(s) \cdot \widetilde{V}_1^1(1, v_1, s)$ and $v_1 = (null, null)$. By construction,  $\widetilde{V}_0$ is an upper confidence bound of $\widetilde{V}_0^1$:
\begin{align*}
    \epsilon_{pe} \ge \widetilde{V}_0 \ge \widetilde{V}_0^1.
\end{align*}
On the other hand, $\sup_{\pi} \PP^{\pi} \left(\exists t_2 > t_1, \ s.t. \ (x_{t_1}, x_{t_2}) \in \mX_l \cap \mX_{l-1}^c \right)$ can be computed through the dynamic programming on $\wQ^*$:
\begin{align*}
    &b_r(i,v,z) = \indic{i=2 \cap z=1 \cap (v' \in \mX_l \cap \mX_{l-1}^c)} \cdot \left( 1 \wedge \sqrt{\frac{\iota_2 }{n_l}} \right), \nonumber \\ 
    &\wQ_t^* ((i,v,s), (a,z)) = b_r + \sum_{s'} T(s'|s,a) \cdot \widetilde{V}_{t+1}^* (i',v',s'), \nonumber
\end{align*}
and
\begin{align*}
    \widetilde{V}_{t}^* (i,v,s) &= \max_{(a,z) \in \widetilde{\mA}} \widetilde{Q}_t^*((i,v,s),(a,z)), & \text{if } t > t_1, \\
    \widetilde{V}_{t}^* (i,v,s) &= \max_{a \in \mA} \widetilde{Q}_t^*((i,v,s),(a,1)), & \text{if } t = t_1, \\
    \widetilde{V}_{t}^* (i,v,s) &= \max_{a \in \mA} \widetilde{Q}_t^*((i,v,s),(a,0)), & \text{if } t < t_1,
\end{align*}
Then the maximum probability of having $(x_{t_1}, x_{t_2}) \in \mX_l \cap \mX_{l-1}^c$ can be computed as the following: 
\begin{align*}
     \widetilde{V}_0^* = \sum_{s} \nu(s) \cdot \widetilde{V}^*(1, v_1, s) = \left(1 \wedge \sqrt{\frac{\iota_2}{n_l}} \right) \cdot \sup_{\pi} \PP^{\pi} \left(\exists t_2 > t_1, \ s.t. \ (x_{t_1}, x_{t_2}) \in \mX_l \cap \mX_{l-1}^c \right).
\end{align*}
Finally, we can inductively show that $\wQ_{t}^1 \ge \wQ_{t}^*$ and $\widetilde{V}_0^1 \ge \widetilde{V}_0^*$ with the setting of confidence interval parameters $\iota_2, \iota_T$. This concludes that
\begin{align*}
    \widetilde{V}_0^1 \ge \widetilde{V}_0^* = \left(1 \wedge \sqrt{\frac{\iota_2}{n_l}} \right) \cdot \sup_{\pi} \PP^{\pi} \left(\exists t_2 > t_1, \ s.t. \ (x_{t_1}, x_{t_2}) \in \mX_l \cap \mX_{l-1}^c \right).
\end{align*}
This concludes that 
\begin{align*}
    \PP_{\Eps_l'}^* \le \sum_{t_1=1}^{H-1} \sup_{\pi} \PP^{\pi} \left(\exists t_2 > t_1, \ s.t. \ (x_{t_1}, x_{t_2}) \in \mX_l \cap \mX_{l-1}^c \right) \le H \epsilon_{pe} \cdot \left(1 \vee \sqrt{n_l / \iota_2} \right).  
\end{align*}

\subsection{Complete Proof of Theorem \ref{theorem:upper_bound}}
\label{appendix:subsec:complete_upper_bound}
In this section, we put the final puzzle piece, a guarantee on total-variation distance between trajectories with different transitions and initial state probabilities, to complete the picture. Let $\mM^1$ be a 2RM-MDP model such that 
\begin{align*}
    &\nu^1(s) = \nu(s), \ T^1(\cdot | s,a) = T(\cdot | s,a), \ R_m^1 (\cdot | s,a) = \hat{R}(\cdot |s,a),  \quad \forall m \in \{1,2\}, s \in \mS, a \in \mA
\end{align*}
For any history-dependent policy $\pi$, we target to bound that
\begin{align*}
    \sum_{\tau} |\PP^{\pi} (\tau) - \hat{\PP}^{\pi} (\tau) | &\le \sum_{\tau} |\PP^{\pi} (\tau) - \PP^{1,\pi} (\tau) | + |\PP^{1,\pi} (\tau) - \hat{\PP}^{\pi} (\tau) |.
\end{align*}
Note that we already have shown that $\sum_{\tau} |\PP^{\pi} (\tau) - \PP^{1,\pi} (\tau) | \le \epsilon / H$ in Section \ref{section:main_analysis} and Appendix \ref{appendix:lemma:basic_tot_bound}. Thus we focus on bounding the second term. We first need the following lemma which is proven in Appendix \ref{appendix:auxiliary_lemmas}.
\begin{lemma}
    \label{lemma:total_variation_occupancy}
    For any history-dependent policy $\pi$, we have
    \begin{align}
        \sum_{\tau} |\PP^{1,\pi} (\tau) - \hat{\PP}^{\pi} (\tau)| &\le \|(\nu - \hat{\nu}) (s_1)\|_1 + \sum_{t=1}^{H-1} \Exs^{1,\pi} \left[\|(T - \hat{T}) (s_{t+1} | s_t, a_t)\|_1\right]. \label{eq:total_occupancy_variation}
    \end{align}
\end{lemma}
After $K$ episodes of pure-exploration, we also get
\begin{align*}
    \|(\nu - \hat{\nu}) (s_1)\|_1 + \sum_{t=1}^{H-1} \Exs^{1,\pi} \left[\|(T - \hat{T}) (s_{t+1} | s_t, a_t)\|_1\right] \le \sqrt{\iota_\nu / K} + \sum_{t=1}^{H-1} \Exs^{1,\pi} \left[\sqrt{\iota_T / n(s_t, a_t)} \right].
\end{align*}

We can show that \eqref{eq:total_occupancy_variation} is at most $2H \widetilde{V_0}$ (recall \eqref{eq:optimistic_value}) from the Bellman-update rule. First we see $\|(\nu - \hat{\nu}) (s_1)\|_1 + \sum_{t=1}^{H-1} \Exs^{1,\pi} \left[\|(T - \hat{T}) (s_{t+1} | s_t, a_t)\|_1\right] \le \widetilde{V}_0^1$ where:
\begin{align*}
    \widetilde{Q}_t^1(s,a) &= \|(T - \hat{T}) (s_{t+1} | s_t, a_t)\|_1 + \sum_{s'} T(s'|s,a) \cdot \widetilde{V}_{t+1}^1(s') \\
    &\le (H+1) \cdot \sqrt{\frac{\iota_T}{n(s_t,a_t)}} + \sum_{s'} \hat{T}(s'|s,a) \cdot \widetilde{V}_{t+1}^1(s'), \\ 
    \widetilde{V}_t^1(s) &= \max_{a} \widetilde{Q}_t^1(s,a), \\
    \widetilde{V}_0^1 &= \|(\nu - \hat{\nu}) (s_1)\|_1 + \sum_{s} \nu(s) \cdot \widetilde{V}_1^1(s) \le (H+1) \cdot \sqrt{\frac{\iota_\nu}{K}} + \sum_{s} \hat{\nu}(s) \cdot \widetilde{V}_1^1(s),
\end{align*}
and $\widetilde{V}_{H+1}^1 = 0$. On the other hand, from the construction of $\widetilde{Q}$ in \eqref{eq:optimistic_value}, for any $i,v,z$, we have $2H \cdot \widetilde{Q}_t ((i,v,s),(a,z)) \ge \widetilde{Q}_t^1 (s,a)$ for all $(s,a)$. Therefore,
\begin{align}
    \label{eq:error_transition_values}
    \sqrt{\iota_\nu / K} + \sum_{t=1}^{H-1} \Exs^{\pi} \left[\sqrt{\iota_T / n(x_t)} \right] \le 2H \widetilde{V}_0 \le 2H \epsilon_{pe}.
\end{align}
With our choice of $\epsilon_{pe} = o(\epsilon / H^3)$, we have $\|\PP^{\pi}(\tau) - \hat{\PP}^{\pi}(\tau)\|_1 \le O(\epsilon/H)$, which then we can conclude that $|V_{\mM}^\pi - V_{\hat{\mM}}^{\pi} | \le H \cdot \|\PP^{\pi}(\tau) - \hat{\PP}^{\pi}(\tau)\|_1 \le O(\epsilon)$.

\section{Proof of the Lower Bound (Theorem \ref{theorem:lower_bound})}
\label{appendix:lower_bound}

We first consider instances of 2RM-MDP with $H = 2, S = 2$ and arbitrary $A \ge 3$. Our construction is as the following:
\begin{enumerate}
    \item Regardless of actions played, both MDPs always start from $s_1$ at $t=1$ and transits to $s_2$ at $t=2$.
    \item At $t = 1$, for all actions $a \in \mA$ except $a_1^*$, both MDPs return a reward sampled from a Bernoulli distribution $Ber(1/2)$.
    \begin{itemize}
        \item $\mM_1$: For the action $a_1^*$, a reward is sampled from $Ber(1/2 + \sqrt{\epsilon})$.
        \item $\mM_2$: For the action $a_1^*$, a reward is sampled from $Ber(1/2 - \sqrt{\epsilon})$.
    \end{itemize}
    \item At $t = 2$, for all actions $a \in \mA$ except $a_{2,1}^*$ and $a_{2,2}^*$, both MDPs return a reward sampled from a Bernoulli distribution $Ber(1/2)$.
    \begin{itemize}
        \item $\mM_1$: For $a_{2,1}^*, a_{2,2}^*$, rewards are sampled from $Ber(1/2 + \sqrt{\epsilon})$ and $Ber(1/2-\sqrt{\epsilon})$ respectively. 
        \item $\mM_2$: For $a_{2,1}^*, a_{2,2}^*$, rewards are sampled from $Ber(1/2 - \sqrt{\epsilon})$ and $Ber(1/2 + \sqrt{\epsilon})$ respectively. 
    \end{itemize}
\end{enumerate}
In this construction, the optimal strategy is to play $a_1^*$ at $t=1$, and depending on the outcome $r_1$, we play either $a_{2,1}^*$ if $r_1 = 1$ or $a_{2,2}^*$ otherwise. By playing this strategy, expected long-term return is
\begin{align*}
    &\frac{1}{2} \sum_{m=1}^2 \PP_m(r_1 = 1) + \PP_m(r_2 = 1 | a_{2,1}^*) \cdot \PP_m(r_1 = 1) + \PP_m(r_2 = 1 | a_{2,2}^*) \cdot \PP_m(r_1 = 0) \\
    &= \frac{1}{2} + \left( (1/2 + \sqrt{\epsilon})^2 + (1/2 - \sqrt{\epsilon})^2 \right) = 1 + 2 \epsilon.
\end{align*}
Note that if we do not play $a_1^*$ at $t=1$, or one of $a_{2,1}^*$ and $a_{2,2}^*$ at $t=2$, then the expected cumulative reward is $1$. Therefore the problem can be reduced to find $a_1^*$ and $a_{2,1}^*$ or $a_{2,2}^*$. 

However, if we play $a_1 \neq a_1^*$ at $t= 1$ or play $a_2 \neq a_{2,1}^*, a_{2,2}^*$ at $t = 2$, {\it i.e.,} if we do not play the right actions at both time steps, then marginal distribution of the sample we get from the environment is always $\PP(r_1, r_2 | a_1, a_2) = \frac{1}{4}$ for all $(r_1, r_2) \in \{0,1\}^{\bigotimes 2}$. Therefore, even if we can access to a full distribution of outcomes from a wrong action sequence, there would be no information gain from playing any wrong sequences of actions, other than removing the played sequence from all $O(A^2)$ possibilities. 

On the other hand, even if we play the correct sequence $(a_1^*,a_{2,1}^*)$ (or $(a_1^*, a_{2,2}^*)$), unless we play it sufficient number of times $O(1/\epsilon^2)$, we cannot distinguish it from other wrong action sequences. That is, the marginal distribution of the reward sequence from the correct action-sequence is close to the one obtained with any wrong action sequence $(a_1, a_2)$ in Kullback-Leibler (KL) divergence:
\begin{align*}
    \sum_{(r_1,r_2) \in \{0,1\}^{\bigotimes 2}} \PP(r_1, r_2 | a_1, a_2) \cdot \log \left(\frac{\PP(r_1, r_2 | a_1, a_2)}{\PP(r_1, r_2 | a_1^*, a_{2,1}^*)} \right) < O(\epsilon^2).
\end{align*}
We can apply a similar lower-bound argument for the multi-armed bandits with $A^2$-arms in \cite{garivier2019explore}, where a general framework for the lower-bound is provided based on information gains from the played policies. Note that there are always $A^2$-possible sequences of actions to play in each episode, and the information gain after each episode is not affected by how the decision in each step is made. This argument gives a $\Omega(A^2 / \epsilon^2)$ lower bound. 

Then following the action-amplification argument (see Appendix A in \cite{kwon2021rl}), we can obtain $\Omega(S^2 A^2 / \epsilon^2)$ lower bound.

\section{Auxiliary Lemmas}
\label{appendix:auxiliary_lemmas}

\subsection{Proof of Lemma \ref{lemma:eps3_parameter_error}}
Recall that $l(x_i) + l(x_j) = \log(\Delta(x_i)) + \log(\Delta(x_j)) = \log |u(x_i, x_j)|$ for all $x_i, x_j \in \mS \times \mA$ when exact model parameters are given from equation \eqref{eq:def_u}. Thus if we have three equations for $(x_1, x_2), (x_2, x_3), (x_3, x_1)$, then we can compute $l(x_1)$ as follows:
\begin{align*}
    l(x_1) = \frac{1}{2} (\log |u(x_1, x_2)| + \log |u(x_2, x_3)| + \log |u(x_3, x_1)|) - \log|u(x_2, x_3)|.
\end{align*}

Now from empirical estimates of the model $\hat{\mM}$, we can compute $\hat{l}(x_1)$ similarly. To simplify the burden in notation, let us denote $z_1 = u(x_1, x_2)$, $z_2 = u (x_2, x_3)$ and $z_3 = u (x_3, x_1)$, and empirical counterparts as $\hat{z}_1, \hat{z}_2, \hat{z}_3$. We first note that
\begin{align*}
    |z_1 - \hat{z}_1| &= |u(x_1, x_2) - \hat{u}(x_1, x_2) | \\
    &\le | \mu(x_1, x_2) - \hat{\mu} (x_1, x_2)| + |p_+(x_1) p_+(x_2) - \hat{p}_+(x_1) \hat{p}_+ (x_2)| + b(x_1, x_2)  \\
    &\le 3 \sqrt{\iota_2 / n(x_1, x_2)} + \epsilon_0^2 \le 0.01 \delta_l \epsilon_l,
\end{align*}
where the last inequality uses $n(x_1, x_2) \ge n_l = C \cdot \iota_2 \delta_l^{-2} \epsilon_l^{-2}$ for a sufficiently large $C > 0$. Similarly, for $z_2 - \hat{z}_2$ and $z_3 - \hat{z}_3$ we can obtain similar inequalities. On the other hand, we note that $|z_1| = \Delta(x_1) \Delta(x_2) \ge \delta_l^2 \gg 0.01 \delta_l \epsilon_l$. Now we can see that
\begin{align*}
    |l(x_1) - \hat{l}(x_1)| &\le \frac{1}{2} \left(\frac{0.01 \delta_l \epsilon_l}{z_1} + \frac{0.01 \delta_l \epsilon_l}{z_2} + \frac{0.01 \delta_l \epsilon_l}{z_3} \right) \\
    &\le \frac{1}{2} \left(\frac{0.01 \delta_l \epsilon_l}{\Delta(x_1) \Delta(x_2)} + \frac{0.01 \delta_l \epsilon_l}{\Delta(x_2) \Delta(x_3)} + \frac{0.01 \delta_l \epsilon_l}{\Delta(x_3) \Delta(x_1)} \right).
\end{align*}
Let $i^* = \arg \max_{i \in [3]} \Delta(x_i) = 2$ (the same argument holds for any $i^*$ by symmetry). First observe that
\begin{align*}
    |l(x_1) - \hat{l}(x_1)| \le \frac{0.01 \epsilon_l}{2} \left( \frac{1}{\Delta(x_1)} + \frac{2}{\Delta(x_2)} \right),
\end{align*}
where we used $\Delta(x_2), \Delta(x_3) \ge \delta_l$. Now converting $l(x)$ to $\Delta(x_1) = \exp(l(x_1))$, we get
\begin{align*}
    \hat{\Delta} (x_1) &\le \Delta(x_1) \exp \left( \frac{0.01 \epsilon_l}{2} \left( \frac{1}{\Delta(x_1)} + \frac{2}{\Delta(x_2)} \right) \right) \\
    &\le \Delta(x_1) \left(1 + 2 \cdot \frac{0.01 \epsilon_l}{2} \left( \frac{1}{\Delta(x_1)} + \frac{2}{\Delta(x_2)} \right) \right) \\
    &\le \Delta(x_1) + 0.03 \epsilon_l,
\end{align*}
where in the second inequality we used $\exp(z) \le 1 + 2z$ for small enough $z$. Similarly, we can also show $\hat{\Delta}(x_1) \ge \Delta(x_1) - 0.03 \epsilon_l$. Symmetric argument holds for $\Delta(x_3)$. For $\Delta(x_2)$, let $i_2^* = \arg \max_{i \in \{1,3\}} \Delta(x_i)$. Then similarly we get
\begin{align*}
    \hat{\Delta}(x_2) &\le \Delta(x_2) \left(1 + \frac{0.01 \epsilon_l}{2} \left( \frac{2}{\Delta(x_2)} + \frac{1}{\Delta(x_{i_2^*})} \right) \right) \\
    &\le \Delta(x_2) + 0.03 \epsilon_l \frac{\Delta(x_2)}{\Delta(x_{i_2^*})}.
\end{align*}
This concludes the proof of Lemma \ref{lemma:eps3_parameter_error}.

\subsection{Proof of Corollary \ref{corollary:big_delta_bound}}
By Lemma \ref{lemma:eps3_parameter_error}, for any $\tau = (x,r)_{1:H} \in \Epst$ and $t \neq t^*$, we have that $\Delta(x_t) \ge \delta_l$ implies $|\Delta(x_t) - \hat{\Delta}(x_t)| \le \epsilon_l / 2$. The bound for $x_{t^*}$ also directly follows from Lemma \ref{lemma:eps3_parameter_error}.

For $\Delta(x_t) < \delta_l$, if $\delta_l = \max(\delta, \epsilon_l) = \epsilon_l$, then we have $\Delta(x_{t^*}) \Delta(x_t) < \epsilon_l \Delta(x_{t^*})$. Now by the constraint \eqref{eq:lp_variables}, we have
\begin{align*}
    |p_- (x_{t^*}) p_-(x_t) - \hat{p}_- (x_{t^*}) \hat{p}_-(x_t)| < 0.01 \delta_l \epsilon_l.
\end{align*}
From this, we have $\hat{\Delta}(x_{t^*}) \hat{\Delta}(x_t) < 1.01 \epsilon_l \Delta(x_{t^*})$. At the same time, by Lemma \ref{lemma:eps3_parameter_error} we also have $\hat{\Delta}(x_{t^*}) \ge 0.5 \Delta(x_t^*)$. Therefore, we get $\hat{\Delta}(x_t) < 3 \epsilon_l$. 

If $\Delta(x_t) < \delta_l$ and $\delta_l = \max(\delta, \epsilon_l) = \delta > \epsilon_l$, then by definition of $\delta$, we have $\Delta(x_t) = 0$. In this case, from $|\hat{\Delta}(x_{t^*}) \hat{\Delta}(x_t) - \Delta(x_{t}) \Delta(x_t)| < \sqrt{\iota_2 / n_l} < 0.03 \epsilon_l \delta$. From this, we get $\hat{\Delta}(x_{t^*}) \hat{\Delta}(x_t)< 0.03 \epsilon_l \delta$. Since $\hat{\Delta}(x_{t^*}) \ge 0.5 \Delta(x_t^*) \ge 0.5\delta$, we get $\hat{\Delta}(x_t) < 2\epsilon_l$.

\subsection{Proof of Lemma \ref{lemma:eps2_count_distinguishable}}
We start with the following inequality:
\begin{align}
    &|u(x_1, x_2) - \hat{u}(x_1, x_2) | \nonumber \\
    &\le | \mu(x_1, x_2) - \hat{\mu} (x_1, x_2)| + |p_+(x_1) p_+(x_2) - \hat{p}_+(x_1) \hat{p}_+ (x_2)| + b(x_1, x_2) \nonumber \\
    &\le 3 \sqrt{\iota_2 / n(x_1, x_2)} + \epsilon_0^2 \le 0.01 \delta_l \epsilon_l  \label{eq:u_diff_concentration}
\end{align}
for any $x_1, x_2$ such that $n(x_1, x_2) \ge n_l$. 

\paragraph {Case I: $\sum_{t=1}^H \indic{\Delta(x_t) \ge \epsilon_l} = 2$.} In this case, let any time-step $t_1$ such that $\hat{\Delta}(x_{t_1}) \ge 2\epsilon_l$, and we show that $\Delta(x_{t_1}) < \epsilon_l$. If it were true, let $t_2, t_3$ be the time steps where $\Delta(x_t) \ge \epsilon_l$ for $t = t_2, t_3$. Then,
\begin{align}
    \label{eq:contradiction_1}
    \hat{\Delta}(x_t) < \Delta(x_t) / 1.5, \qquad t = t_2, t_3.
\end{align}
If this is the case, then $|\hat{u}(x_{t_2}, x_{t_3}) - u(x_{t_2}, x_{t_3})| > u(x_{t_2}, x_{t_3}) / 2 > \delta_l \epsilon_l / 2$. However, visit counts for all pairs in a same trajectory satisfies $n(x_{t_2}, x_{t_3}) \ge C \cdot \iota_2 \delta_l^{-2} \epsilon_l^{-2}$, and thus $|\hat{u}(x_{t_2}, x_{t_3}) - u(x_{t_2}, x_{t_3})| < 0.01 \delta_l \epsilon_l$, which is contradiction (see equation \eqref{eq:u_diff_concentration}). 

Now we argue that \eqref{eq:contradiction_1} is true. Suppose for $t_2$, we have $\hat{\Delta}(x_{t_2}) \ge \Delta(x_{t_2}) / 1.5$. Then, 
\begin{align*}
    \hat{u} (x_{t_1}, x_{t_2}) = \hat{\Delta}(x_{t_1}) \hat{\Delta}(x_{t_2}) > \frac{4}{3} \epsilon_l \Delta(x_{t_2}). 
\end{align*}
On the other hand, we have $u(x_{t_1}, x_{t_2}) < \epsilon_l \Delta(x_{t_2})$. Now we can see that
\begin{align*}
    |\hat{u} (x_{t_1}, x_{t_2}) - u(x_{t_1}, x_{t_2}) | \ge \frac{1}{3} \epsilon_l \Delta(x_{t_2}) > \frac{1}{3} \epsilon_l \delta_l,
\end{align*}
which is again contradiction since $|\hat{u}(x_{t_2}, x_{t_3}) - u(x_{t_2}, x_{t_3})| < 0.01 \delta_l \epsilon_l$. Hence we showed that $\hat{\Delta}(x_t) > 2\epsilon_l$ only if $\Delta(x_t) \ge \epsilon_l$. This ensures that $\sum_{t=1}^H \indic{\Delta(x_t) \ge \epsilon_l \cup \hat{\Delta}(x_t) \ge 2\epsilon_l} = 2$.

\paragraph {Case II: $\sum_{t=1}^H \indic{\Delta(x_t) \ge \epsilon_l} \le 1$.} Here, we show that there should not exist any two $t_1, t_2$ such that $\hat{\Delta}(x_{t_1}), \hat{\Delta}(x_{t_2}) \ge 2\epsilon_l$ and at the same time $\Delta(x_{t_1}), \Delta(x_{t_2}) < \epsilon_l$. If this is the case, then we immediately get $\sum_{t=1}^H \indic{\Delta(x_t) \ge \epsilon_l \cup \hat{\Delta}(x_t) \ge 2\epsilon_l} \le 2$.

Now we show that there exist no such pair of $t_1, t_2$. We first consider when $\delta_l = \epsilon_l$. In this case, we have
\begin{align*}
    |\hat{u}(x_{t_1}, x_{t_2}) - u(x_{t_1}, x_{t_2})| \ge 3\epsilon_l^2 = 3 \delta_l \epsilon_l,
\end{align*}
which is a contradiction to equation \eqref{eq:u_diff_concentration}. 

In the other case $\delta_l = \max(\delta, \epsilon_l) > \epsilon_l$, note that this is equivalent to $\sum_{t=1}^H \indic{\Delta(x_t) > 0} \le 1$. We show that whenever $\Delta(x_1) = 0$, then the LP \eqref{eq:LP_formulation} returns a solution such that $\hat{\Delta}(x_1) \le \epsilon_0^2$. Note that whenever $\Delta(x_1) = 0$, then for any $x_2$, we have $u(x_1, x_2) = \Delta(x_1) \Delta(x_2) = 0$. This implies, with high probability, for any $x_2$ we have
\begin{align*}
    c_u (x_1, x_2) < 0 < c_l (x_1, x_2). 
\end{align*}
Hence there is no lower bound constraint \eqref{eq:constrinats_lower_bound} for $\hat{l}(x_1) + \hat{l}(x_2)$ as $c_u(x_1, x_2) \cdot c_l(x_1, x_2) < 0$, and $\hat{l}(x_1)$ can always take the minimum possible value from \eqref{eq:constraints_regularity}, which gives $\hat{l}(x_1) = 10 \log(\epsilon_0)$. Hence by the minimizing objective in \eqref{eq:LP_formulation}, we get $\hat{\Delta}(x_1) \le \epsilon_0^2$. This proves that whenever $\hat{\Delta}(x) \ge 2\epsilon_l$, we must have $\Delta(x) > 0$, which proves $\sum_{t=1}^H \indic{\Delta(s_t, a_t) \ge \epsilon_l \cup \hat{\Delta}(s_t, a_t) \ge 2\epsilon_l} \le 2$.

\subsection{Proof of Lemma \ref{lemma:surrogate_models_proximity}}
For the true model $\mM$ and $\mM^1$, we can see that
\begin{align*}
    |R_m^1(r|x) - R_m(r|x)| &= |\epsilon_l - (R_+(r|x) - R_m(r|x)) - 2\epsilon_l R_+ (r|x)|, \\
    &\le \epsilon_l |1 - 2R_+(r|x)| + 2\epsilon_l \le 3\epsilon_l, & \text { if } \Delta(x) < 2\epsilon_l, \\
    |R_m^1(r|x) - R_m(r|x)| &= |\epsilon_l - 2 \epsilon_l R_m(r|x)| \le 3\epsilon_l, & \text { if } \Delta(x) \ge 2\epsilon_l,
\end{align*}
For the empirical model $\hat{\mM}$ and $\mM^2$, we first note that 
\begin{align*}
    |\hat{R}_+ (r|x) - R_+(r|x)| \le \sqrt{\iota_2 / n(x)} \le \sqrt{\iota_2 / n_l} \le 0.01 \delta_l \epsilon_l.
\end{align*} 
for all $x: \exists (x_1, x_2) \in \mX_l, \text{ s.t. } x = x_1 \text{ or } x_2$. Then, we can check that
\begin{align*}
    |R_m^2(r|x) - \hat{R}_m(r|x)| &= |\epsilon_l - (R_+(r|x) - \hat{R}_m(r|x)) - 2\epsilon_l R_+ (r|x)|, \\
    &\le \epsilon_l |1 - 2R_+(r|x)| + |R_+(r|x) - \hat{R}_+(r|x)| + |\hat{R}_+(r|x) - \hat{R}_m(r|x)| \\
    &\le 4\epsilon_l, 
\end{align*}
if $\Delta(x) < 2\epsilon_l$, and for the else case, for both $m = 1,2$ we have
\begin{align*}
    |R_m^2(r|x) - R_m(r|x)| &= \epsilon_l|1 - 2R_+(r|x)| + |R_+(r|x) - \hat{R}_+(r|x)| \le 3\epsilon_l.
\end{align*}

\subsection{Proof of Lemma \ref{lemma:total_variation_occupancy}}
This is a special case of Lemma B.3 in \cite{kwon2021rl} with same transition and initial probability parameters for all contexts, as RM-MDP can be also considered as a special case of LMDP. To make the paper self-contained, we reproduce the proof. For any $t \in [H]$ and $m \in [2]$, we start with
\begin{align*}
    \sum_{(s,a,r)_{1:t}} &|\PP_m^{1,\pi} - \hat{\PP}_m^{\pi}| ((s,a,r)_{1:t}) =  \sum_{(s,a,r)_{1:t-1}, s_t} |\PP_m^{1, \pi} - \hat{\PP}_m^{\pi}| ((s,a,r)_{1,t-1}, s_t) \\
    &\le \sum_{(s,a,r)_{1:t-1}, s_t} |\PP_m^{1, \pi} - \hat{\PP}_m^{\pi}| ((s,a,r)_{1:t-1}) \hat{T} (s_t | s_{t-1}, a_{t-1}) \\
    &\quad + \sum_{(s,a,r)_{1:t-1}, s_t} \PP_m^{1, \pi} ((s,a,r)_{1:t-1}) |T - \hat{T}| (s_{t} | s_{t-1}, a_{t-1}) \\
    &\le \sum_{(s,a,r)_{1:t-1}} |\PP_m^{1,\pi} - \hat{\PP}_m^{\pi}| ((s,a,r)_{1:t-1}) \\
    &\quad + \sum_{(s,a,r)_{1:t-1}} \|(T - \hat{T}) (s_t | s_{t-1}, a_{t-1})\|_1 \PP_m^{1,\pi} ((s,a,r)_{1:t-1}),
\end{align*}
where in the first equality follows since we use the same reward models for $\mM^1$ and $\hat{\mM}$. Recursively applying the same argument, we get
\begin{align*}
    \sum_{(s,a,r)_{1:H}} &|\PP_m^{1,\pi} - \hat{\PP}_m^{\pi}| ((s,a,r)_{1:H}) \\
    &\le \|(\nu - \hat{\nu}) (s_1)\|_1 + \sum_{t=2}^H \sum_{(s,a,r)_{1:t-1}} \|(T - \hat{T}) (s_t | s_{t-1}, a_{t-1})\|_1 \PP_m^{1,\pi} ((s,a,r)_{1:t-1}).
\end{align*}
Finally, we observe that
\begin{align*}
    \sum_{(s,a,r)_{1:H}} &|\PP^{1,\pi} - \hat{\PP}^{\pi}| ((s,a,r)_{1:H}) = \frac{1}{2} \sum_{m=1}^2 \sum_{(s,a,r)_{1:H}} |\PP_m^{1,\pi} - \hat{\PP}_m^{\pi}| ((s,a,r)_{1:H}) \\
    &\le \|(\nu - \hat{\nu}) (s_1)\|_1 + \sum_{t=2}^H \sum_{(s,a,r)_{1:t-1}} \|(T - \hat{T}) (s_t | s_{t-1}, a_{t-1})\|_1 \PP^{1,\pi} ((s,a,r)_{1:t-1}) \\
    &= \|(\nu - \hat{\nu}) (s_1)\|_1 + \Exs^{1,\pi} \left[\|(T - \hat{T}) (s_t | s_{t-1}, a_{t-1})\|_1 \right],
\end{align*}
which concludes the lemma.

\section{Towards Non-Uniform Mixing Weights}
\label{appendix:sign_ambiguity}
In this appendix, we discuss the high-level idea of how to handle non-uniform mixing weights $w_1 = w, w_2 = 1-w$. For a simplified discussion, let us assume $w_1, w_2 = \Omega(1)$, $\delta = \Omega (1)$ and the true $T, \nu$ are known in advance. We first recheck several quantities to check what becomes different. First, now the average reward is $p_+(x) := w\cdot p_1(x) + (1-w) \cdot p_2(x)$. Let differences in rewards be $p_-(x) := p_1(x) - p_2(x)$. For any $x_i, x_j \in \mS \times \mA$, the reward correlation function now becomes (compare this to \eqref{eq:def_mu}):
\begin{align}
    \mu(x_i, x_j) = w \cdot p_1(x_i) p_1(x_j) + (1-w) \cdot p_2(x_i) p_2(x_j). \label{eq:def_mu_non_uniform}
\end{align}
If we construct a matrix $B \in \mathbb{R}^{SA \times SA}$ indexed by state-actions such that at its $(i,j)$ entry is given by:
\begin{align*}
    B_{i,j} = \mu(x_i, x_j) - p_+(x_i) p_+(x_j), 
\end{align*}
then we get $B = w (1-w) \cdot q q^\top$ where $q$ is a vector indexed by $x$ such that $q_i = p_-(x_i)$. Note that the only difference from $w = 1/2$ case is the overall re-scaling factor $w (1-w)$. This suggests that we can still extract the same information on reward differences (compare this to \eqref{eq:def_u})
$$u(x_i, x_j) := w(1-w) \cdot p_-(x_i) p_-(x_j) = \mu(x_i,x_j) - p_+(x_i) p_+(x_j).$$
This means the overall algorithm remains the same: we can still first get correlations from pure-exploration (Algorithm \ref{algo:pure_explore}), and then recovers magnitude of $|p_-(x_i)|$ (Algorithm \ref{algo:LMDP_recovery_detail}).

What complicates the problem is the recovery of signs: after solving a 2-SAT problem to find all signs of $\hat{p}_-(x)$ that satisfy \eqref{eq:constraints_sign}, still there remains an ambiguity whether we have $sign(\hat{p}_-(x)) = sign(p_-(x))$ or $sign(\hat{p}_-(x)) = -sign(p_-(x))$ (with pair-wise consistency to satisfy \eqref{eq:constraints_sign}), because either $sign(p_-(x))$ or $-sign(p_-(x))$ is a consistent solution for \eqref{eq:constraints_sign}. This is not a problem when the model is symmetric in signs of $p_-(x)$ (recall the symmetry argument in Appendix \ref{appendix:equivalence_in_signs}), but when the prior is non-uniform, then we also need to find the exact signs of $p_-(x)$. To see this, observe that
\begin{align*}
    p_1(x) = p_+(x) + (1-w) \cdot p_-(x), \ \ p_2(x) = p_+(x) - w \cdot p_-(x),
\end{align*}
and that changing the sign of $p_-(x)$ may result in a different set of pairs $(p_1(x), p_2(x))$.

Now we elaborate how to resolve the sign-ambiguity issue if $w \neq 1/2$. Let us consider several disjoint subsets of state-actions $G_1, G_2, ..., G_q \subseteq \mS \times \mA$ such that the following holds: (a) $\cup_{l=1}^q G_l = \{x \in \mS \times \mA : |p_-(x)| \ge \delta \}$, and (b) any $x_i, x_j \in \mS \times \mA$ belongs to the same $G_l$ for some $l \in [q]$ if one decides $sign(x_i)$ then $sign(x_j)$ is automatically decided. One can construct $\{G_l\}_{l=1}^q$ when we formulate the sign assignment problem to a 2-SAT problem. Now, if $q = O(1)$,  then there will be only a small number of possible candidate empirical models that satisfy \eqref{eq:constraints_sign}. If that is the case, we can compute the optimal policy for each candidate, run each optimal policy on the real environment, and take the one with more (estimated) long-term rewards. 

In general, $q$ can be at most $O(SA)$, and thus creating too many candidate models from the combination of consistent sign assignments. In such case, we can extract information from third order correlations. Observe that, for some $(x_i, x_j, x_k) \in (\mS\times\mA)^{\bigotimes 3}$, we define:
\begin{align*}
    \mu(x_i, x_j, x_k) := \Exs[r_i r_j r_k | x_i, x_j, x_k ] = w \cdot p_1(x_i) p_1(x_j) p_1(x_k) + (1-w) \cdot p_2(x_i) p_2(x_j) p_2(x_k).
\end{align*}
Some algebra shows that from this, we can extract a quantity
\begin{align*}
    u(x_i, x_j, x_k) &:= \mu(x_i, x_j, x_k) - p_+(x_i) p_+(x_j) p_+(x_k) \\
    &\quad - \left(p_+(x_i) u(x_j, x_k) + p_+(x_j) u(x_i, x_k) + p_+(x_k) u(x_i, x_j) \right) \\
    &= w(1-w) (1-2w) \cdot p_-(x_i) p_-(x_j) p_-(x_k).
\end{align*}
Suppose that all three state-actions belong to the same group, {\it i.e.,} $x_i, x_j, x_k \in G_l$ for some $l \in [q]$. If $|1 - 2w|$ is not too small (say, larger than $\epsilon / H^2$), then by looking at the sign of $u(x_i, x_j, x_k)$, we can decide the signs of not only $x_i, x_j, x_k$, but also all elements in $G_l$. 

Now note that, for each $l \in [q]$, there are only two possible assignments of signs for all elements in $G_l$. Pick any element $x_l \in G_l$. Non-negligible $u(x_i, x_j, x_k)$ suggests that we can perform a hypothesis-testing whether $sign(x_l) = +1$ or $-1$, because joint-probability of $(r(x_i), r(x_j), r(x_k))$ for any $x_i, x_j, x_k \in G_l$, for each case are at least apart by $|w(1-w)(1-2w) \delta^{3}|$ in total-variation distance. Thus if we can collect $O(|w(1-w)(1-2w) \delta^{3}|^{-2})$ number of samples of any third-order correlations from $G_l$, we can decide $sign(x_l)$. While we do not investigate this direction further in detail, it will be of independent interest to obtain a tight sample complexity (both upper and lower bounds) for non-uniform priors with an optimal design of recovery mechanism for $sign(x_l)$. We leave it as future work.

%% file: main.bbl
\begin{thebibliography}{10}

\bibitem{asmuth2009bayesian}
J.~Asmuth, L.~Li, M.~L. Littman, A.~Nouri, and D.~Wingate.
\newblock A bayesian sampling approach to exploration in reinforcement
  learning.
\newblock In {\em Proceedings of the Twenty-Fifth Conference on Uncertainty in
  Artificial Intelligence}, pages 19--26, 2009.

\bibitem{aspvall1979linear}
B.~Aspvall, M.~F. Plass, and R.~E. Tarjan.
\newblock A linear-time algorithm for testing the truth of certain quantified
  boolean formulas.
\newblock {\em Information Processing Letters}, 8(3):121--123, 1979.

\bibitem{azar2017minimax}
M.~G. Azar, I.~Osband, and R.~Munos.
\newblock Minimax regret bounds for reinforcement learning.
\newblock In {\em International Conference on Machine Learning}, pages
  263--272. PMLR, 2017.

\bibitem{azizzadenesheli2016reinforcement}
K.~Azizzadenesheli, A.~Lazaric, and A.~Anandkumar.
\newblock Reinforcement learning of {POMDP}s using spectral methods.
\newblock In {\em Conference on Learning Theory}, pages 193--256, 2016.

\bibitem{boots2011closing}
B.~Boots, S.~M. Siddiqi, and G.~J. Gordon.
\newblock Closing the learning-planning loop with predictive state
  representations.
\newblock {\em The International Journal of Robotics Research}, 30(7):954--966,
  2011.

\bibitem{brunskill2013sample}
E.~Brunskill and L.~Li.
\newblock Sample complexity of multi-task reinforcement learning.
\newblock In {\em Uncertainty in Artificial Intelligence}, page 122. Citeseer,
  2013.

\bibitem{buchholz2019computation}
P.~Buchholz and D.~Scheftelowitsch.
\newblock Computation of weighted sums of rewards for concurrent {MDP}s.
\newblock {\em Mathematical Methods of Operations Research}, 89(1):1--42, 2019.

\bibitem{chades2012momdps}
I.~Chad{\`e}s, J.~Carwardine, T.~Martin, S.~Nicol, R.~Sabbadin, and O.~Buffet.
\newblock Momdps: a solution for modelling adaptive management problems.
\newblock In {\em Twenty-Sixth AAAI Conference on Artificial Intelligence
  (AAAI-12)}, 2012.

\bibitem{chan2013learning}
S.-O. Chan, I.~Diakonikolas, X.~Sun, and R.~A. Servedio.
\newblock Learning mixtures of structured distributions over discrete domains.
\newblock In {\em Proceedings of the twenty-fourth annual ACM-SIAM symposium on
  Discrete algorithms}, pages 1380--1394. SIAM, 2013.

\bibitem{chen2017convex}
Y.~Chen, X.~Yi, and C.~Caramanis.
\newblock Convex and nonconvex formulations for mixed regression with two
  components: Minimax optimal rates.
\newblock {\em IEEE Transactions on Information Theory}, 64(3):1738--1766,
  2017.

\bibitem{cohen2021solving}
M.~B. Cohen, Y.~T. Lee, and Z.~Song.
\newblock Solving linear programs in the current matrix multiplication time.
\newblock {\em Journal of the ACM (JACM)}, 68(1):1--39, 2021.

\bibitem{dann2018oracle}
C.~Dann, N.~Jiang, A.~Krishnamurthy, A.~Agarwal, J.~Langford, and R.~E.
  Schapire.
\newblock On oracle-efficient {PAC} {RL} with rich observations.
\newblock In {\em Advances in neural information processing systems}, pages
  1422--1432, 2018.

\bibitem{dasgupta1999learning}
S.~Dasgupta.
\newblock Learning mixtures of gaussians.
\newblock In {\em 40th Annual Symposium on Foundations of Computer Science
  (Cat. No. 99CB37039)}, pages 634--644. IEEE, 1999.

\bibitem{de1989mixtures}
R.~D. De~Veaux.
\newblock Mixtures of linear regressions.
\newblock {\em Computational Statistics \& Data Analysis}, 8(3):227--245, 1989.

\bibitem{diakonikolas2019robust}
I.~Diakonikolas, G.~Kamath, D.~Kane, J.~Li, A.~Moitra, and A.~Stewart.
\newblock Robust estimators in high-dimensions without the computational
  intractability.
\newblock {\em SIAM Journal on Computing}, 48(2):742--864, 2019.

\bibitem{du2019provably}
S.~Du, A.~Krishnamurthy, N.~Jiang, A.~Agarwal, M.~Dudik, and J.~Langford.
\newblock Provably efficient rl with rich observations via latent state
  decoding.
\newblock In {\em International Conference on Machine Learning}, pages
  1665--1674, 2019.

\bibitem{du2019good}
S.~S. Du, S.~M. Kakade, R.~Wang, and L.~F. Yang.
\newblock Is a good representation sufficient for sample efficient
  reinforcement learning?
\newblock In {\em International Conference on Learning Representations}, 2019.

\bibitem{efroni2019tight}
Y.~Efroni, N.~Merlis, M.~Ghavamzadeh, and S.~Mannor.
\newblock Tight regret bounds for model-based reinforcement learning with
  greedy policies.
\newblock In {\em Advances in Neural Information Processing Systems}, pages
  12224--12234, 2019.

\bibitem{feldman2008learning}
J.~Feldman, R.~O'Donnell, and R.~A. Servedio.
\newblock Learning mixtures of product distributions over discrete domains.
\newblock {\em SIAM Journal on Computing}, 37(5):1536--1564, 2008.

\bibitem{freund1999estimating}
Y.~Freund and Y.~Mansour.
\newblock Estimating a mixture of two product distributions.
\newblock In {\em Proceedings of the twelfth annual conference on Computational
  learning theory}, pages 53--62, 1999.

\bibitem{garivier2019explore}
A.~Garivier, P.~M{\'e}nard, and G.~Stoltz.
\newblock Explore first, exploit next: The true shape of regret in bandit
  problems.
\newblock {\em Mathematics of Operations Research}, 44(2):377--399, 2019.

\bibitem{guo2016pac}
Z.~D. Guo, S.~Doroudi, and E.~Brunskill.
\newblock A {PAC} {RL} algorithm for episodic {POMDP}s.
\newblock In {\em Artificial Intelligence and Statistics}, pages 510--518,
  2016.

\bibitem{hallak2015contextual}
A.~Hallak, D.~Di~Castro, and S.~Mannor.
\newblock Contextual markov decision processes.
\newblock {\em arXiv preprint arXiv:1502.02259}, 2015.

\bibitem{hoffmann2020learning}
J.~Hoffmann, S.~Basu, S.~Goel, and C.~Caramanis.
\newblock Learning mixtures of graphs from epidemic cascades.
\newblock In {\em International Conference on Machine Learning}, pages
  4342--4352. PMLR, 2020.

\bibitem{hsu2012spectral}
D.~Hsu, S.~M. Kakade, and T.~Zhang.
\newblock A spectral algorithm for learning hidden markov models.
\newblock {\em Journal of Computer and System Sciences}, 78(5):1460--1480,
  2012.

\bibitem{jain2014learning}
P.~Jain and S.~Oh.
\newblock Learning mixtures of discrete product distributions using spectral
  decompositions.
\newblock In {\em Conference on Learning Theory}, pages 824--856. PMLR, 2014.

\bibitem{jaksch2010near}
T.~Jaksch, R.~Ortner, and P.~Auer.
\newblock Near-optimal regret bounds for reinforcement learning.
\newblock {\em Journal of Machine Learning Research}, 11:1563--1600, 2010.

\bibitem{jiang2017contextual}
N.~Jiang, A.~Krishnamurthy, A.~Agarwal, J.~Langford, and R.~E. Schapire.
\newblock Contextual decision processes with low bellman rank are
  pac-learnable.
\newblock In {\em International Conference on Machine Learning}, pages
  1704--1713. PMLR, 2017.

\bibitem{jin2019learning}
C.~Jin, T.~Jin, H.~Luo, S.~Sra, and T.~Yu.
\newblock Learning adversarial mdps with bandit feedback and unknown
  transition.
\newblock {\em arXiv preprint arXiv:1912.01192}, 2019.

\bibitem{jin2020sample}
C.~Jin, S.~M. Kakade, A.~Krishnamurthy, and Q.~Liu.
\newblock Sample-efficient reinforcement learning of undercomplete {POMDP}s.
\newblock {\em arXiv preprint arXiv:2006.12484}, 2020.

\bibitem{jin2020reward}
C.~Jin, A.~Krishnamurthy, M.~Simchowitz, and T.~Yu.
\newblock Reward-free exploration for reinforcement learning.
\newblock {\em arXiv preprint arXiv:2002.02794}, 2020.

\bibitem{jin2020provably}
C.~Jin, Z.~Yang, Z.~Wang, and M.~I. Jordan.
\newblock Provably efficient reinforcement learning with linear function
  approximation.
\newblock In {\em Conference on Learning Theory}, pages 2137--2143. PMLR, 2020.

\bibitem{kaufmann2021adaptive}
E.~Kaufmann, P.~M{\'e}nard, O.~D. Domingues, A.~Jonsson, E.~Leurent, and
  M.~Valko.
\newblock Adaptive reward-free exploration.
\newblock In {\em Algorithmic Learning Theory}, pages 865--891. PMLR, 2021.

\bibitem{kearns2002near}
M.~Kearns and S.~Singh.
\newblock Near-optimal reinforcement learning in polynomial time.
\newblock {\em Machine learning}, 49(2):209--232, 2002.

\bibitem{krishnamurthy2016pac}
A.~Krishnamurthy, A.~Agarwal, and J.~Langford.
\newblock {PAC} reinforcement learning with rich observations.
\newblock In {\em Advances in Neural Information Processing Systems}, pages
  1840--1848, 2016.

\bibitem{kwon2021rl}
J.~Kwon, Y.~Efroni, C.~Caramanis, and S.~Mannor.
\newblock {RL} for latent {MDPs}: Regret guarantees and a lower bound.
\newblock {\em arXiv preprint arXiv:2102.04939}, 2021.

\bibitem{kwon2020minimax}
J.~Kwon, N.~Ho, and C.~Caramanis.
\newblock On the minimax optimality of the {EM} algorithm for learning
  two-component mixed linear regression.
\newblock {\em arXiv preprint arXiv:2006.02601}, 2020.

\bibitem{maillard2014latent}
O.-A. Maillard and S.~Mannor.
\newblock Latent bandits.
\newblock In {\em International Conference on Machine Learning}, pages
  136--144, 2014.

\bibitem{neu2012adversarial}
G.~Neu, A.~Gyorgy, and C.~Szepesv{\'a}ri.
\newblock The adversarial stochastic shortest path problem with unknown
  transition probabilities.
\newblock In {\em Artificial Intelligence and Statistics}, pages 805--813.
  PMLR, 2012.

\bibitem{osband2014model}
I.~Osband and B.~Van~Roy.
\newblock Model-based reinforcement learning and the eluder dimension.
\newblock In {\em Advances in Neural Information Processing Systems}, pages
  1466--1474, 2014.

\bibitem{osband2017posterior}
I.~Osband and B.~Van~Roy.
\newblock Why is posterior sampling better than optimism for reinforcement
  learning?
\newblock In {\em International Conference on Machine Learning}, pages
  2701--2710. PMLR, 2017.

\bibitem{pineau2006anytime}
J.~Pineau, G.~Gordon, and S.~Thrun.
\newblock Anytime point-based approximations for large {POMDP}s.
\newblock {\em Journal of Artificial Intelligence Research}, 27:335--380, 2006.

\bibitem{russo2013eluder}
D.~Russo and B.~Van~Roy.
\newblock Eluder dimension and the sample complexity of optimistic exploration.
\newblock {\em Advances in Neural Information Processing Systems}, 2013.

\bibitem{simchowitz2019non}
M.~Simchowitz and K.~G. Jamieson.
\newblock Non-asymptotic gap-dependent regret bounds for tabular mdps.
\newblock In {\em Advances in Neural Information Processing Systems}, pages
  1153--1162, 2019.

\bibitem{steimle2018multi}
L.~N. Steimle, D.~L. Kaufman, and B.~T. Denton.
\newblock Multi-model markov decision processes.
\newblock {\em Optimization Online URL http://www. optimization-online.
  org/DB\_FILE/2018/01/6434. pdf}, 2018.

\bibitem{sutton2018reinforcement}
R.~S. Sutton and A.~G. Barto.
\newblock {\em Reinforcement learning: An introduction}.
\newblock MIT press, 2018.

\bibitem{tossou2019near}
A.~Tossou, D.~Basu, and C.~Dimitrakakis.
\newblock Near-optimal optimistic reinforcement learning using empirical
  bernstein inequalities.
\newblock {\em arXiv preprint arXiv:1905.12425}, 2019.

\bibitem{wainwright2019high}
M.~J. Wainwright.
\newblock {\em High-dimensional statistics: A non-asymptotic viewpoint},
  volume~48.
\newblock Cambridge University Press, 2019.

\bibitem{yu2009markov}
J.~Y. Yu, S.~Mannor, and N.~Shimkin.
\newblock Markov decision processes with arbitrary reward processes.
\newblock {\em Mathematics of Operations Research}, 34(3):737--757, 2009.

\bibitem{zanette2019tighter}
A.~Zanette and E.~Brunskill.
\newblock Tighter problem-dependent regret bounds in reinforcement learning
  without domain knowledge using value function bounds.
\newblock In {\em International Conference on Machine Learning}, pages
  7304--7312. PMLR, 2019.

\end{thebibliography}
